\documentclass[10pt]{article} 
\PassOptionsToPackage{table}{xcolor} 
\usepackage[accepted]{tmlr}

\usepackage{amsmath,amsfonts,bm}

\def\eqref#1{equation~\ref{#1}}

\def\1{\bm{1}}

\DeclareMathAlphabet{\mathsfit}{\encodingdefault}{\sfdefault}{m}{sl}
\SetMathAlphabet{\mathsfit}{bold}{\encodingdefault}{\sfdefault}{bx}{n}

\usepackage{hyperref}
\usepackage{url}

\usepackage{graphicx}
\usepackage{subfig}
\usepackage{afterpage}
\usepackage{wrapfig} 

\usepackage{enumitem}
\usepackage[most]{tcolorbox}
\usepackage{algorithm}
\usepackage{algpseudocode}
\usepackage{caption} 

\usepackage[table]{xcolor} 
\usepackage{booktabs}
\usepackage{multirow}
\usepackage{makecell}

\usepackage{enumitem}

\title{UniDoc: Towards Unified Thinking in Documents}

\author{\name Qixin Xu$^{*}$ \email xqx23@mails.tsinghua.edu.cn \\
      \addr Tsinghua University
      \AND
      \name Haozhe Wang$^{*}$ \email jasper.whz@outlook.com \\
      \addr Hong Kong University of Science and Technology
      \AND
      \name Che Liu \email che.liu21@imperial.ac.uk \\
      \addr Imperial College London
      \AND
      \name Fangzhen Lin \email flin@cse.ust.hk \\
      \addr Hong Kong University of Science and Technology
      \AND
      \name Wenhu Chen \email hustchenwenhu@gmail.com \\
      \addr University of Waterloo}

\def\month{11}  
\def\year{2026} 
\def\openreview{\url{https://openreview.net/forum?id=c4HyrJIiEa}} 

\begin{document}

\maketitle
{\renewcommand{\thefootnote}{\fnsymbol{footnote}}
\footnotetext[1]{Equal contribution.}}

\begin{abstract}
Answering questions over long, visually rich documents demands both cross-page structural reasoning and fine-grained visual understanding. Existing approaches sacrifice one for the other: retrieval-augmented methods fragment documents and lose inter-page structure; monolithic vision-language models preserve structure but scale poorly beyond a few pages. This impasse stems from a false premise: that the macroscopic task of evidence retrieval and the microscopic task of detailed reasoning must either be split into separate systems or be processed identically. By decoupling their perceptual resolutions rather than their architectures, a single model can achieve the structural holism of monolithic processing alongside the efficiency of targeted retrieval. UniDoc realizes this through two resolution-adaptive stages: a Fast Reading stage processes all pages at low resolution to structurally identify evidence pages, and a Focused Thinking stage re-processes only those pages at high resolution for multimodal inference. To train this pipeline, we construct 30K multi-hop reasoning instances with stage-level supervision. On MMLongBench-Doc, UniDoc yields relative accuracy gains of 55.7\% on Qwen2.5-VL-7B and 16.9\% on Qwen3-VL-8B, while significantly reducing token cost relative to uniform high-resolution processing. Error analysis identifies page retrieval and fine-grained visual reasoning as the remaining bottlenecks, delineating concrete targets for future work. Code and data are available at \url{https://github.com/QixinXu/CogDoc}.
\end{abstract}

\section{Introduction}
\label{sec:intro}

\begin{wrapfigure}{R}{0.52\linewidth}
  \vspace{-0.8\baselineskip}
  \centering
  \includegraphics[width=0.9\linewidth]{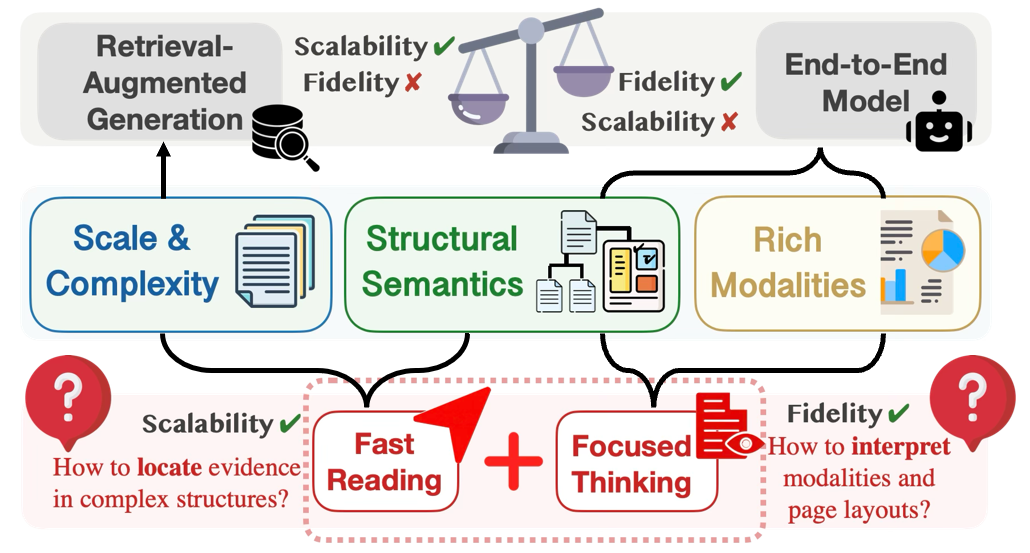}
  \caption{Existing paradigms face a fundamental scalability--fidelity trade-off. UniDoc addresses this by jointly optimizing evidence retrieval and multimodal reasoning within a single model through two resolution-adaptive stages: Fast Reading and Focused Thinking.}
  \label{fig:teaser}
  \vspace{-0.8\baselineskip}
\end{wrapfigure}

Real-world tasks such as auditing financial statements by cross-referencing tables with textual disclosures, synthesizing hypotheses across scientific corpora, and connecting evidentiary threads in legal proceedings demand systems that can navigate hundreds of visually rich pages while performing precise evidence grounding and multimodal reasoning. No current system provides both at scale. This gap stems from three intertwined challenges. \textbf{Rich Modalities}: text, tables, charts, and diagrams whose true meaning often emerges only when they are jointly interpreted. \textbf{Structural Semantics}: 2D spatial layout, heading hierarchies, and typographic cues that encode meaning beyond textual content. \textbf{Scale and Complexity}: documents spanning hundreds of pages that necessitate multi-hop reasoning across distant evidence.

These difficulties create a tension that current paradigms fail to resolve. The dominant approach, retrieval-augmented generation (RAG)~\citep{lewis2021retrievalaugmentedgenerationknowledgeintensivenlp, shi2023replugretrievalaugmentedblackboxlanguage, cho2024m3docragmultimodalretrievalneed, xia2024mmed, ma2024unifyingmultimodalretrievaldocument, faysse2025colpaliefficientdocumentretrieval, yu2025visragvisionbasedretrievalaugmentedgeneration}, addresses scale by fragmenting documents into chunks and retrieving subsets via local semantic similarity. However, this fragmentation is architecturally at odds with fidelity. Retrieval based on local similarity cannot capture structural semantics that span page boundaries, such as inferring subordination from heading hierarchy across sections. Chunking further severs the links between modalities, dissociating a figure from its referencing text. 

Conversely, monolithic end-to-end vision-language models (VLMs)~\citep{kim2022ocrfreedocumentunderstandingtransformer, ye2023ureaderuniversalocrfreevisuallysituated, hu2024mplugdocowl15unifiedstructure} preserve fidelity by processing full visual pages, but cannot efficiently handle the hundreds of pages common in practice. The field thus faces an apparent impasse: scalability through fragmentation or fidelity through monolithic processing.

This impasse stems from a shared design limitation: treating evidence retrieval and multimodal reasoning as either architecturally separate systems (RAG) or as an undifferentiated, monolithic computation (end-to-end VLMs). We argue that this dichotomy is a false premise. Our key insight is that, rather than severing the architecture (like RAG) or homogenizing the computation (like VLMs), \textbf{a unified model can resolve this tension by decoupling the perceptual resolutions applied to each task.}

Concretely, locating evidence pages demands broad, structural understanding of the full document at low perceptual cost; extracting answers from those pages demands fine-grained, high-resolution multimodal reasoning over a narrow context. This asymmetry admits a natural sequential decomposition \emph{within a single model}. Both stages are reasoning over the same document at different resolutions, so one model can perform both, unlike RAG, where a frozen retriever is decoupled from the reader.

Therefore, we introduce \textbf{UniDoc}, a single-model unified training framework that operationalizes this asymmetry. In the first stage, \textit{Fast Reading}, the model ingests the entire document at low resolution and performs structural reasoning. It parses heading hierarchies, identifies section boundaries, and follows cross-references to locate the sparse set of pages where evidence resides. In the second stage, \textit{Focused Thinking}, the same model re-processes only the localized pages at high resolution for fine-grained multimodal inference: cross-referencing text with table cells, interpreting chart trends alongside surrounding narrative, and synthesizing information across the retrieved evidence. Because Stage~1 compresses the search space before Stage~2 demands perceptual fidelity, UniDoc recovers high-resolution perception without paying the token cost of processing every page at full resolution. No external retrieval modules or document fragmentation are needed.

Realizing this framework requires addressing two key challenges. First, the opposing demands of broad abstraction versus pixel-level scrutiny make joint training non-trivial; we systematically investigate post-training strategies to resolve this, contrasting supervised fine-tuning followed by reinforcement learning (SFT+RL) against pure reinforcement learning (RL)~\citep{shao2024deepseekmathpushinglimitsmathematical, wang2025vl, xu2026prismpromotingsamplingdiversity}. Second, existing document reasoning datasets lack the multi-page compositional queries and stage-level supervision that this paradigm demands. We bridge this gap by contributing a curated dataset of 30K high-quality multimodal document instances, produced by a principled synthesis engine that generates complex, multi-hop queries requiring evidence from multiple pages, paired with explicit annotations for both the retrieval and reasoning stages.

Pairing this high-quality data with our unified framework yields large and consistent gains across backbones. On Qwen2.5-VL-7B, UniDoc improves accuracy on MMLongBench-Doc by 11.8 points and on SlideVQA by 19.4 points; on the stronger Qwen3-VL-8B, the gains persist (+7.6 on MMLongBench-Doc), indicating that explicit decomposition of retrieval and reasoning is not subsumed by stronger backbone capacity (Table~\ref{tab:main_results}). These gains come at lower token cost than monolithic full-resolution processing. Beyond benchmark numbers, the two-stage decomposition provides a diagnostic lens that monolithic models and fragmented pipelines cannot offer. Aligning errors with stages reveals that structural understanding for evidence retrieval remains the dominant bottleneck. As retrieval improves, fine-grained multimodal reasoning emerges as the second major failure mode, charting two concrete directions for future work.

\begin{figure*}[t]
  \centering

  \includegraphics[width=0.98\textwidth]{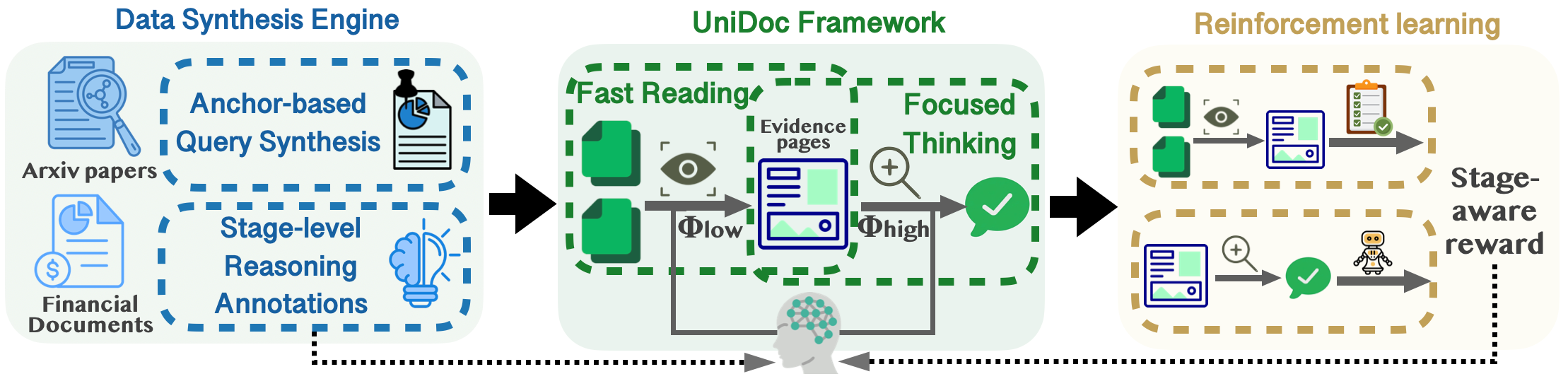}
  
  \caption{Overview of our methodology. The Data Synthesis Engine produces complex queries and stage-level annotations from raw documents. UniDoc processes them through Fast Reading and Focused Thinking stages, optimized end-to-end via Reinforcement Learning with stage-aware rewards.}
  \label{fig:my_large_figure}
\end{figure*}

Our contributions are fourfold:
\begin{itemize}[itemsep=2pt, topsep=2pt, parsep=0pt, partopsep=0pt]
\item We introduce UniDoc, a resolution-adaptive framework that decomposes document understanding into structural evidence retrieval (Fast Reading) and fine-grained multimodal inference (Focused Thinking). This substantially relaxes the scalability--fidelity tension without external modules.
\item We contribute 30K multi-hop reasoning instances with stage-level supervision, with which we optimize UniDoc via stage-aware RL and systematically investigate post-training strategies.
\item UniDoc delivers large empirical gains across two distinct backbones, achieving higher accuracy while substantially reducing input-token cost relative to monolithic full-resolution processing.
\item The two-stage decomposition yields a diagnostic error taxonomy that delineates structural retrieval and fine-grained multimodal reasoning as the two remaining dominant failure modes.
\end{itemize}

\section{Related Work}
\label{sec:formatting}
\subsection{Document Understanding}
Early document understanding focused on page-level fidelity with layout-aware pre-training models~\citep{Xu_2020, huang2022layoutlmv3pretrainingdocumentai, kim2022ocrfreedocumentunderstandingtransformer, tang2023unifyingvisiontextlayout}, which jointly model text and 2D position. While these monolithic models excel at handling \textit{Rich Modalities} and \textit{Structural Semantics}~\citep{appalaraju2021docformerendtoendtransformerdocument, davis2022endtoenddocumentrecognitionunderstanding}, their computational cost prohibits scaling to lengthy documents. To circumvent this scalability bottleneck, the dominant paradigm shifted to fragmentation-based pipelines, notably Retrieval-Augmented Generation (RAG)~\citep{lewis2021retrievalaugmentedgenerationknowledgeintensivenlp, shi2023replugretrievalaugmentedblackboxlanguage, cho2024m3docragmultimodalretrievalneed}. This approach evolved from text-centric methods~\citep{karpukhin2020densepassageretrievalopendomain, sachan2023improvingpassageretrievalzeroshot} that rely on lossy OCR, to more recent Visual-RAG frameworks~\citep{ma2024unifyingmultimodalretrievaldocument, yu2025visragvisionbasedretrievalaugmentedgeneration, wang2026pixelrag} that encode multimodal embeddings directly from visual chunks. Between these two extremes, GRAM~\citep{blau2024gramglobalreasoningmultipage} adds global cross-page reasoning layers, and CREAM~\citep{zhang2024cream} adopts a coarse-to-fine retrieval pipeline, but both keep localization and answering separately parameterized.

However, both RAG variants are fundamentally constrained by their reliance on local query-content similarity~\citep{asai2023selfraglearningretrievegenerate, gao2024retrievalaugmentedgenerationlargelanguage}. By physically fragmenting the document, this mechanism is inherently blind to the document's holistic topology~\citep{sarthi2024raptorrecursiveabstractiveprocessing} and severs the complex interdependencies between pages~\citep{wang2023docllmlayoutawaregenerativelanguage}, thus fundamentally failing at tasks that require multi-hop, compositional reasoning~\citep{han2025retrievalaugmentedgenerationgraphsgraphrag, cheng2025dualragdualprocessapproachintegrate}.

Concurrently, agentic frameworks~\citep{yu2025visualdocumentunderstandingquestion, wang2025vidoragvisualdocumentretrievalaugmented, han2025mdocagentmultimodalmultiagentframework, zheng2026docvcoarsetofineinteractivevisualreasoning, xia2025mmedagent} attempt to recover this multi-step reasoning capability through dynamic task decomposition. Yet, they rely on iterative, external LLM calls and complex routing, introducing prohibitive computational overhead that limits practical scalability.
None of these paradigms resolves the scalability--fidelity tension without significant compromises.

\subsection{Multimodal Reasoning}

In parallel, the field of multimodal reasoning has seen a surge of innovations for single-image tasks. Notable advancements encompass multimodal chain-of-thought reasoning~\citep{wei2023chainofthoughtpromptingelicitsreasoning, xu2025llavacotletvisionlanguage, wang2025multimodalchainofthoughtreasoningcomprehensive}, active perception, iterative visual deliberation such as ``thinking with images'', and fine-grained grounding via tool use~\citep{wang2025pixelreasonerincentivizingpixelspace, openai2025thinkingwithimages,fu2025refocusvisualeditingchain, zhang2025perl, xia2025agent0}. These paradigms transform perception from a passive encoding into a dynamic process and have proven highly effective at enabling high-resolution, fine-grained visual analysis within a local, single-image context.

However, this powerful, fine-grained reasoning paradigm has thus far been confined to single-image or few-image contexts~\citep{lu2024mathvistaevaluatingmathematicalreasoning,yue2023mmmu,zhang2025pixelcraft}, with long-context reasoning remaining largely unexplored. In the video domain, recent
works~\citep{fan2024videoagentmemoryaugmentedmultimodalagent, fu2025lover1advancinglongvideo, meng2025openo3videogroundedvideo} have begun to address this by selecting a subset of
keyframes before performing detailed reasoning on the selected
frames. Yet video frames are temporally continuous and highly
redundant, making frame selection a forgiving perceptual filtering
task. Document pages, by contrast, are semantically
discrete, each carrying unique, non-redundant content, and omitting a
single page can irreversibly lose critical evidence. Furthermore, navigating a lengthy document requires parsing logical topology (e.g., cross-references, heading hierarchies) rather than relying on visual saliency cues prevalent in videos, presenting a fundamentally structure-driven localization challenge that existing frame-selection methods do not address. A complementary line of work compresses textual content into a smaller number of visual tokens to extend the effective context~\citep{wei2025deepseekocrcontextsopticalcompression}, which likewise treats perceptual resolution as a budget to be allocated rather than a fixed property of the input.


\section{Methods}
Our approach is built on a simple observation: not all pages of a document require the same perceptual resolution. Identifying \emph{where} evidence resides is a structural reasoning task that can be performed at low perceptual cost over the full document; understanding \emph{what} that evidence means is a multimodal inference task that demands high-resolution perception over only a few pages. We exploit this asymmetry to avoid the two extremes: monolithic high-resolution processing (computationally prohibitive and infeasible for RL) and RAG-style fragmentation (destructive to structural semantics). We operationalize this \textbf{resolution asymmetry} through two contributions. The \textit{Resolution-Adaptive Two-Stage Reasoning Framework} (Section~\ref{sec:framework}) decomposes document reasoning into sequential low-resolution localization and high-resolution inference within a single, jointly trained model. The \textit{Anchor-Grounded Data Synthesis Engine} (Section~\ref{sec: data engine}) produces the multi-page, cross-modal training data this framework requires.

\subsection{Resolution-Adaptive Two-Stage Reasoning}
\label{sec:framework}

\textbf{Stage~1: Fast Reading -- Structure-Aware Evidence Localization.}
Unlike conventional retrieval, which matches queries to chunks via context-blind embedding similarity, Fast Reading performs \emph{structural reasoning} over the document's architecture. Given a query $q$ and a document $d$ comprising $N$ pages, the model (parameterized by shared weights $\theta$) ingests $d$ at low resolution $\phi_{\text{low}}$. It generates a reasoning trajectory $\tau_1$ that leverages heading hierarchies, section boundaries, and cross-references to identify a compact evidence set $\mathcal{P}_{\text{ev}} \subset d$, where $|\mathcal{P}_{\text{ev}}| = k \ll N$:
\begin{equation*}
  \tau_1 = \bigl(a_1^{(1)}, \dots, a_{T_1}^{(1)}\bigr)
  \sim \pi_{\theta,\, \phi_{\text{low}}}\!\bigl(\cdot \mid q,\; d\bigr),
\end{equation*}
where the terminal action $a_{T_1}^{(1)}$ outputs the page identifiers $\mathcal{P}_{\text{ev}}$. This stage addresses the \textit{Scale and Complexity} and \textit{Structural Semantics} challenges: the model reasons about the document's global organization, rather than local token overlap, to locate evidence that may be distributed across distant sections.

\noindent\textbf{Stage~2: Focused Thinking -- High-Resolution Multimodal Inference.}
The \textit{same} model $\theta$ re-processes only the localized pages $\mathcal{P}_{\text{ev}}$ at high resolution $\phi_{\text{high}}$, now performing fine-grained multimodal reasoning. It cross-references table cells with textual claims, interprets chart trends alongside surrounding text, and synthesizes information across heterogeneous layout elements to produce the final answer $\hat{y}$:
\begin{equation*}
  \tau_2 = \bigl(a_1^{(2)}, \dots, a_{T_2}^{(2)}\bigr)
  \sim \pi_{\theta,\, \phi_{\text{high}}}\!\bigl(\cdot \mid q,\; \mathcal{P}_{\text{ev}}\bigr),
\end{equation*}
where $a_{T_2}^{(2)} = \hat{y}$. This stage addresses the \textit{Rich Modalities} and \textit{Structural Semantics} challenges: with the search space compressed by Stage~1, the model can allocate its full perceptual capacity to the compositional and structure-aware multimodal inference that document understanding demands. Because Stage~2 observes only $\mathcal{P}_{\text{ev}}$, it cannot recover a page omitted by Stage~1; Appendix~\ref{app:foveated} evaluates a variant that retains non-selected pages at $\phi_{\text{low}}$.

Concretely, we set $\phi_{\text{low}}$ to $262144$ and $\phi_{\text{high}}$ to $1204224$ maximum pixels per page, using the same budgets during RL and at inference. The two are constrained differently: $\phi_{\text{low}}$ is bounded by context length, since Fast Reading ingests all $N$ pages in a single pass, whereas $\phi_{\text{high}}$ applies only to the $k \ll N$ localized pages and is free to scale. Section~\ref{sec:efficiency} traces the effect of $\phi_{\text{high}}$ on the accuracy--cost frontier.

\textbf{A single shared-parameter model.}
Both stages are executed by a single, jointly optimized model $\theta$. Unlike RAG pipelines that physically decouple the retriever and reader, UniDoc executes both stages with one set of parameters; Appendix~\ref{app:decoupled} compares this against a decoupled pipeline of two independently trained specialists. Crucially, no gradient flows from Stage~2 outcomes back to Stage~1 decisions; each stage is optimized independently against its own stage-level supervision. Performance gains can transfer implicitly through the shared weights. This design trades complex cross-stage credit assignment for training efficiency, a choice enabled by the precise ground-truth page labels from our data engine.

\textbf{Joint optimization with stage-aware rewards.}
Each stage demands a distinct correctness metric, necessitating a \textit{stage-aware hybrid reward}: rule-based set matching for Fast Reading and LLM-as-a-Judge~\citep{zheng2023judgingllmasajudgemtbenchchatbot} semantic evaluation for Focused Thinking. We optimize $\pi_\theta$ via reinforcement learning on a joint dataset $\mathcal{D}_{\text{RL}} = \mathcal{D}_{\text{FR}} \cup \mathcal{D}_{\text{FT}}$, where $\mathcal{D}_{\text{FR}}$ contains Stage~1 queries and $\mathcal{D}_{\text{FT}}$ contains Stage~2 queries paired with their ground-truth evidence pages $\mathcal{P}_{\text{gt}}$, sampled jointly within each batch. Training follows the GRPO framework~\citep{shao2024deepseekmathpushinglimitsmathematical}, with full details in Appendix~\ref{appd: reward design}.

Groups for advantage computation are formed strictly within each stage: the $G$ sampled trajectories for a Stage~1 query are evaluated via rule-based matching, while those for a Stage~2 query are evaluated via LLM judgment. Crucially, no cross-stage mixing occurs during advantage normalization, ensuring each trajectory is compared only against others from the same stage.

\subsection{Anchor-Grounded Data Synthesis}
\label{sec: data engine}

The two-stage framework imposes specific data requirements that existing document QA benchmarks do not satisfy. Current datasets predominantly feature single-page lookup queries that exercise neither the broad structural reasoning needed for Fast Reading nor the fine-grained multimodal reasoning demanded by Focused Thinking~\citep{mathew2021docvqadatasetvqadocument, mathew2021infographicvqa, masry2022chartqabenchmarkquestionanswering}. We bridge this gap with a synthesis engine that produces 30K high-quality training instances, pairing complex multi-page queries with explicit stage-level reasoning annotations.

\subsubsection{Anchor-Grounded Query Synthesis}
\label{sec:data synthesis}

Our key observation is that multimodal elements (figures, tables, and charts) serve as natural \textit{information anchors} within documents: they concentrate rich visual semantics at specific locations while being referenced by textual passages that may reside on entirely different pages. This anchor-reference structure is precisely what makes document reasoning challenging. We exploit it to synthesize queries that, \textit{by construction}, mandate both structural localization and cross-modal synthesis.

Concretely, for each anchor $z$ in a document $d$ (e.g., ``Figure~5'' on page~8), we extract its visual content $I_z$ and retrieve all textual passages $T_z$ that reference it (e.g., paragraphs on pages~3 and~20 that cite ``Figure~5''). A query $q_z$ is then generated by a VLM conditioned on both $I_z$ and $T_z$, yielding questions designed to require cross-page, cross-modal synthesis. This anchor-based design simultaneously creates training signal for both stages:

\begin{itemize}[leftmargin=*, itemsep=2pt, topsep=2pt, parsep=0pt, partopsep=0pt]
    \item \textbf{Fast Reading signal:} Because the anchor $I_z$ and its referencing text $T_z$ typically reside on different pages, the ground-truth evidence set $\mathcal{P}_{\text{gt}}$ is inherently distributed across the document (e.g., $\mathcal{P}_{\text{gt}} = \{3, 8, 20\}$). Locating all relevant pages requires the model to reason about the document's structural organization by following cross-references and understanding section hierarchy, rather than relying on surface-level similarity.

    \item \textbf{Focused Thinking signal:} Because every query is grounded in a multimodal anchor, answering requires fine-grained multimodal reasoning: extracting trends from the visual elements in $I_z$, connecting them to quantitative claims in $T_z$, and synthesizing a coherent answer. This directly exercises fine-grained multimodal inference.
\end{itemize}

The synthesis procedure, including anchor detection, reference retrieval, query generation, and quality filtering, is formalized in Algorithm~\ref{alg:synthesis}.

\begin{algorithm}[t]
\centering
\begin{minipage}{0.9\linewidth}
\footnotesize
\linespread{0.9}\selectfont
\caption{Anchor-Grounded Query Synthesis}
\label{alg:synthesis}
\begin{algorithmic}[1]
\State \textbf{Input:} Document corpus $\mathcal{D}$, Anchor detection $\mathcal{G}$, Reference retrieval $\mathcal{R}$, VLM generator $\mathcal{M}$, Quality filter $\mathcal{F}$
\State \textbf{Output:} Query set $\mathcal{Q}_{\text{syn}}$
\vspace{2pt}
\State $\mathcal{Q}_{\text{syn}} \leftarrow \emptyset$
\ForAll{document $d \in \mathcal{D}$}
    \State $\text{Anchors} \leftarrow \mathcal{G}(d)$ \Comment{Detect figures, tables, charts}
    \ForAll{anchor $z \in \text{Anchors}$}
        \State $I_z \leftarrow \text{Crop}(d, \text{BBox}(z))$ \Comment{Extract visual anchor}
        \State $n_z \leftarrow \text{ExtractName}(\text{Caption}(z))$ \Comment{e.g., ``Figure~5''}
        \State $T_z \leftarrow \mathcal{R}(d, n_z)$ \Comment{Retrieve referencing passages}
        \State $q_{\text{cand}} \leftarrow \mathcal{M}(I_z, T_z)$ \Comment{Generate query from anchor + text}
        \If{$\mathcal{F}(q_{\text{cand}})$} \Comment{Verify answerability}
            \State $\mathcal{Q}_{\text{syn}} \leftarrow \mathcal{Q}_{\text{syn}} \cup \{q_{\text{cand}}\}$
        \EndIf
    \EndFor
\EndFor
\State \textbf{return} $\mathcal{Q}_{\text{syn}}$
\end{algorithmic}
\end{minipage}
\end{algorithm}

\subsubsection{Stage-Level Reasoning Annotations}
\label{sec:thinking_patterns}

The two-stage decomposition raises a natural question: can the model discover effective reasoning strategies for both stages through RL exploration alone, or is explicit supervision necessary to guide optimization in such a vast search space? To investigate this, we synthesize chain-of-thought annotations tailored to the distinct demands of each stage.

\textbf{Fast Reading annotations} provide supervision for \textit{structure-aware evidence localization}. Rather than simply listing ground-truth pages, we use a powerful VLM to reverse-engineer reasoning trajectories: given the target evidence pages, the VLM reconstructs the structural reasoning steps required to arrive at them, such as following cross-references (e.g., ``as shown in Figure~5'') and leveraging heading hierarchies. These annotations train the model to perform strategic document navigation rather than brute-force page scanning.

\textbf{Focused Thinking annotations} provide supervision for \textit{adaptive multimodal reasoning}. We observe that the primary bottleneck in this stage is not information extraction per se, but the \textit{layout grounding} step -- deciding which visual and textual elements on a page contain task-relevant information. The complexity of this grounding varies significantly with query difficulty and visual density. Accordingly, we synthesize annotations at multiple granularity levels: concise for straightforward lookups and elaborated for queries requiring multi-step cross-modal synthesis. This heterogeneous supervision encourages the model to calibrate its reasoning depth to query complexity, explicitly addressing the lack of adaptive computation recently identified as a key bottleneck in reasoning models~\citep{tang2026ringzeroscalingzerorl}.

Because the two stages impose different optimization pressures (global structural abstraction versus local perceptual scrutiny), jointly training both within a single model may introduce inter-stage interference. We empirically investigate this tension in Section~\ref{sec: analysis}, comparing direct RL, SFT followed by RL, and stage-separated training.

\definecolor{myblue}{rgb}{0.0, 0.25, 1.0}
\newcommand{\deltavalue}[3]{\hspace{#3mm}#1\scalebox{0.7}{\textcolor{myblue}{{+#2}}}}
\newcommand{\minusdeltavalue}[3]{\hspace{#3mm}#1\scalebox{0.7}{\textcolor{red}{{-#2}}}}
\newcommand{\conf}[1]{\scalebox{0.7}{\textcolor{gray}{(\textit{#1})}}}

\section{Experiments}

\subsection{Experimental Setup}

\textbf{Training data.}
The primary source of our training data is our Data Synthesis Engine, detailed in Section~\ref{sec:data synthesis} and Appendix~\ref{app:corpus}. 
To improve domain diversity, we supplement this synthesized data with the established training sets of DUDE~\citep{vanlandeghem2023documentunderstandingdatasetevaluation} and SlideVQA~\citep{tanaka2023slidevqadatasetdocumentvisual}. We also process these queries through our synthesis pipeline to generate ``stage-level reasoning annotations'' detailed in Section \ref{sec:thinking_patterns}.

\smallskip
\textbf{Models.}
We build UniDoc on two backbones of different generations: Qwen2.5-VL-7B-Instruct~\citep{bai2025qwen25vltechnicalreport} and the natively long-context Qwen3-VL-8B~\citep{bai2025qwen3vltechnicalreport}, yielding UniDoc-7B and UniDoc-8B. Main results (Table~\ref{tab:main_results}) and the efficiency analysis (Section~\ref{sec:efficiency}) cover both models; unless otherwise noted, all ablations and analyses use UniDoc-7B, while UniDoc-8B confirms that our training recipe generalizes to a newer backbone.

\smallskip
\textbf{Benchmarks.} 
Our evaluation consists of four challenging, multi-page document understanding benchmarks, selected to test diverse reasoning capabilities. We use \textbf{MMLongBench-Doc}~\citep{ma2024mmlongbenchdocbenchmarkinglongcontextdocument} to test long-context compositional reasoning (avg. 47.5 pages); \textbf{DUDE}~\citep{vanlandeghem2023documentunderstandingdatasetevaluation} to evaluate fine-grained information extraction from forms; \textbf{SlideVQA}~\citep{tanaka2023slidevqadatasetdocumentvisual} to challenge the understanding of complex layouts and charts in presentations; and \textbf{MP-DocVQA}~\citep{tito2023hierarchicalmultimodaltransformersmultipage} to evaluate cross-page navigation for localizing answers. A detailed description of each benchmark is provided in Appendix~\ref{appd:benchmarks}. 

\begin{table*}[t]
\caption{Quantitative experimental results. Size refers to the number of parameters. Paradigm refers to the specific method to derive the answer. Reported metrics include ANLS, Accuracy, and F1 score. Best performances among open-weight models are in \textbf{bold}, second best are \underline{underlined}. MP-DocVQA is evaluated as a held-out benchmark and marked with $^{\dagger}$: its training set is excluded from our training data, while all other columns are trained on. Bolding and underlining are therefore applied only to the trained columns. Results with MP-DocVQA included in training are reported in Appendix~\ref{app:mpdocvqa}. Dash (-) indicates the metric was not originally evaluated for that dataset.}
\label{tab:main_results}
    \centering
    {
    \resizebox{\textwidth}{!}{
    \begin{tabular}{l r c c cc cc c}
    \toprule
\multirow{3}{*}{\textbf{Model}} & \multirow{3}{*}{\textbf{Size}} & \multirow{3}{*}{\textbf{Paradigm}} & \multirow{2}{*}{\textbf{DUDE}} & \multicolumn{2}{c}{\textbf{MMLong}} & \multicolumn{2}{c}{\multirow{2}{*}{\textbf{SlideVQA}}} & {\textbf{MP-}} \\
& & & & \multicolumn{2}{c}{\textbf{bench-doc}} & & & \textbf{DocVQA}$^{\dagger}$ \\
\cmidrule(r){5-6} \cmidrule(r){7-8}
& & & ANLS & Acc. & F1 & Acc. & F1 & Acc. \\
\midrule
Average Images Per Question & - & - & 5.7 & \multicolumn{2}{c}{47.5} & \multicolumn{2}{c}{20.0} & 8.3 \\
\midrule
GPT-4o mini~\citep{hurst2024gpt} & - & End-to-End & 46.5 & 28.6 & 29.4 & 48.1 & 60.7 & - \\
GPT-4o~\citep{hurst2024gpt} & - & End-to-End & 52.7 & 41.6 & 42.3 & 53.1 & 65.8 & - \\
GPT-4.1~\citep{openai2025gpt41} & - & End-to-End & 50.2 & 45.6 & 49.7 & 61.5 & 74.7 & - \\
Claude-3.7-Sonnet & - & End-to-End & 58.1 & 33.9 & 38.4 & 62.9 & 76.3 & - \\
\midrule
Qwen2.5-VL-7B-Instruct~\citep{bai2025qwen25vltechnicalreport} & 7B & End-to-End & 41.8 & 21.2 & 20.7 & 38.9 & 49.8 & 57.3 \\
Qwen2.5-VL-72B-Instruct~\citep{bai2025qwen25vltechnicalreport} & 72B & End-to-End & 44.5 & 24.9 & 24.6 & 38.8 & 54.4 & 64.2 \\
MiMo-VL-7B-SFT~\citep{coreteam2025mimovltechnicalreport} & 7B & End-to-End & 43.1 & 27.5 & 27.2 & 48.0 & 61.9 & 63.9 \\
InternVL3-8B~\citep{zhu2025internvl3exploringadvancedtraining} & 8B & End-to-End & 43.4 & 29.6 & 28.8 & 51.6 & 64.4 & 83.3 \\
Qwen3-VL-8B~\citep{bai2025qwen3vltechnicalreport} & 8B & End-to-End & 45.4 & \underline{44.9} & \underline{43.7} & 46.3 & 57.9 & 72.8 \\
\midrule
M3DocRAG~\citep{cho2024m3docragmultimodalretrievalneed}~\conf{arXiv'24} & 10B & RAG & 39.5 & 21.0 & 22.6 & 48.8 & 57.9 & 84.4 \\
mPLUG-DocOwl2~\citep{hu2024mplugdocowl2highresolutioncompressingocrfree}~\conf{ACL'25} & 8B & End-to-End & 45.6 & 13.4 & 8.9 & 25.4 & 30.4 & 69.4 \\
VisRAG~\citep{yu2025visragvisionbasedretrievalaugmentedgeneration}~\conf{ICLR'25} & 8B & RAG & 43.1 & 18.8 & 18.3 & 50.9 & 52.4 & - \\
SV-RAG~\citep{chen2025svragloracontextualizingadaptationmllms}~\conf{ICLR'25} & 8B & RAG & 45.0 & 23.0 & 24.2 & 32.0 & 34.3 & 71.0 \\
Docopilot~\citep{duan2025docopilot}~\conf{CVPR'25} & 8B & End-to-End & 40.7 & 28.8 & 23.0 & 35.7 & 43.1 & 81.3 \\
VDocRAG~\citep{tanaka2025vdocragretrievalaugmentedgenerationvisuallyrich}~\conf{CVPR'25} & 8B & RAG & 44.0 & - & - & 42.0 & 44.0 & - \\
FRAG~\citep{huang2025fragframeselectionaugmented}~\conf{arXiv'25} & 8B & RAG & 43.1 & 31.3 & 30.7 & 52.4 & 63.9 & 77.8 \\
\midrule
UniDoc-7B & 7B & Unified Thinking & \hspace{4mm}\deltavalue{\underline{46.2}}{4.4}{0} & \hspace{5mm}\deltavalue{33.0}{11.8}{0} & \hspace{5mm}\deltavalue{31.5}{10.8}{0} & \hspace{5mm}\deltavalue{\underline{58.3}}{19.4}{0} & \hspace{5mm}\deltavalue{\underline{67.9}}{18.1}{0} & \hspace{5mm}\deltavalue{75.0}{17.7}{0} \\
UniDoc-8B & 8B & Unified Thinking & \hspace{4mm}\deltavalue{\textbf{47.8}}{2.4}{0} & \hspace{5mm}\deltavalue{\textbf{52.5}}{7.6}{0} & \hspace{5mm}\deltavalue{\textbf{51.0}}{7.3}{0} & \hspace{5mm}\deltavalue{\textbf{65.1}}{18.8}{0} & \hspace{5mm}\deltavalue{\textbf{73.1}}{15.2}{0} & \hspace{5mm}\deltavalue{81.4}{8.6}{0} \\
\bottomrule
\end{tabular}
}
}
\end{table*}

\smallskip

\textbf{Baselines.} We compare our method against three categories of state-of-the-art baselines. First, as proprietary and general-purpose reference points, we include GPT-4.1~\citep{openai2025gpt41}, GPT-4o, GPT-4o mini~\citep{hurst2024gpt}, Claude-3.7-Sonnet, Qwen2.5-VL~\citep{bai2025qwen25vltechnicalreport}, Qwen3-VL~\citep{bai2025qwen3vltechnicalreport}, InternVL~3~\citep{zhu2025internvl3exploringadvancedtraining}, and MiMo-VL~\citep{coreteam2025mimovltechnicalreport}. Second, we evaluate document-understanding specialists spanning two dominant paradigms: multimodal Retrieval-Augmented Generation (RAG) and holistic end-to-end frameworks. The RAG baselines encompass standard models (M3DocRAG~\citep{cho2024m3docragmultimodalretrievalneed}, VisRAG~\citep{yu2025visragvisionbasedretrievalaugmentedgeneration}) alongside advanced variants utilizing self-retrieval adapters (SV-RAG~\citep{chen2025svragloracontextualizingadaptationmllms}), self-supervised pre-training (VDocRAG~\citep{tanaka2025vdocragretrievalaugmentedgenerationvisuallyrich}), and efficient page selection (FRAG~\citep{huang2025fragframeselectionaugmented}). For the efficiency analysis in Table~\ref{tab:wallclock}, we additionally include ColPali-RAG~\citep{faysse2025colpaliefficientdocumentretrieval}, a representative embedding-based visual retriever that pairs ColPali page retrieval with the same Qwen backbone, serving as a strong latency-oriented RAG reference. Finally, the end-to-end specialists include DocOwl2~\citep{hu2024mplugdocowl2highresolutioncompressingocrfree} and Docopilot~\citep{duan2025docopilot}, which avoid fragmentation via high-resolution compression and native document-level dependency learning, respectively.

\subsection{Main Results}
\label{sec: main results}
We present a comprehensive evaluation across multiple benchmarks in Table~\ref{tab:main_results}. We apply our framework to two backbones, Qwen2.5-VL-7B-Instruct and natively long-context Qwen3-VL-8B, each optimized via direct Reinforcement Learning (Direct RL), yielding UniDoc-7B and UniDoc-8B.

The framework produces large and consistent gains on both backbones. On Qwen2.5-VL-7B, it improves accuracy on MMLongBench-Doc by 11.8 points and on SlideVQA by 19.4 points, and ANLS on the fine-grained DUDE benchmark by 4.4 points. These gains persist on stronger backbones. Qwen3-VL-8B is a more capable starting point: its native long-context ability reaches 44.9 on MMLongBench-Doc, above UniDoc-7B. Yet our framework still adds +7.6 there and +18.8 on SlideVQA. The benefit of decomposing structural retrieval from high-resolution reasoning is therefore not subsumed by stronger native long-context capacity.

These gains also allow both models to outperform larger and proprietary systems, which we treat as reference points rather than controlled comparisons. Despite its size, UniDoc-7B exceeds the $10\times$ larger Qwen2.5-VL-72B-Instruct on both MMLongBench-Doc and DUDE, and outperforms GPT-4o on SlideVQA accuracy. UniDoc-8B reaches 52.5 on MMLongBench-Doc, surpassing the proprietary references GPT-4o, GPT-4.1, and Claude-3.7-Sonnet on this long-context benchmark.

Finally, to test generalization to a distribution unseen during training, we evaluate on MP-DocVQA, whose training set is excluded from our data. While these results are not directly comparable to specialists fine-tuned specifically on MP-DocVQA, UniDoc improves over each base model (57.3$\rightarrow$75.0 on Qwen2.5-VL-7B; 72.8$\rightarrow$81.4 on Qwen3-VL-8B), indicating robust transfer to unseen document types and query distributions.

Table~\ref{tab:wallclock} reports overall wall-clock latency and token consumption alongside accuracy. While UniDoc requires more inference time than embedding-based RAG (9.4s vs. 5.3s), it achieves higher end-to-end accuracy and halves the total token consumption (26.3k vs. 53.5k).

\begin{table}[h]
\centering
\caption{Comparison of accuracy and computational cost on MMLongBench-Doc. Wall-clock time includes all preprocessing and inference stages.}
\label{tab:wallclock}
\begin{tabular}{lccc}
\toprule
\textbf{Method} & \textbf{Acc.} & \textbf{Tokens/Doc} & \textbf{Wall-Clock (s/doc)} \\
\midrule
Full high-res (all pages) &  21.2 & 49.8k & 11.0 \\
ColPali-RAG~\citep{faysse2025colpaliefficientdocumentretrieval} & 31.9 & 47.4k\,$+$\,6.1k & 5.3 \\
UniDoc-7B & \textbf{33.0} & \textbf{26.3k} & 9.4 \\
\bottomrule
\end{tabular}
\end{table}

\begin{table*}[h]
\centering
\small 
\caption{Ablation study of our core components. We quantify the impact of each component by comparing UniDoc-7B against the Base Model and three ablated variants, each removing one of our core contributions.}
\label{tab:main_ablation}
\resizebox{\textwidth}{!}{
\begin{tabular}{l|ccc|cccc}
\toprule
 & \multicolumn{3}{c|}{\textbf{Our Components}} & \multicolumn{4}{c}{\textbf{Benchmarks}} \\
\cmidrule(r){2-4} \cmidrule(l){5-8}
\textbf{Model Configuration} & \begin{tabular}[c]{@{}c@{}}Two-Stage\\ Framework\end{tabular} & \begin{tabular}[c]{@{}c@{}}Data Engine\\ (Synthesized)\end{tabular} & \begin{tabular}[c]{@{}c@{}}RL Training\\ (Direct RL)\end{tabular} & \begin{tabular}[c]{@{}c@{}}DUDE\\ (ANLS)\end{tabular} & \begin{tabular}[c]{@{}c@{}}MMLong.\\ (Acc.)\end{tabular} & \begin{tabular}[c]{@{}c@{}}SlideVQA\\ (Acc.)\end{tabular} & \begin{tabular}[c]{@{}c@{}}MP-DocVQA\\ (Acc.)\end{tabular} \\
\midrule
Base Model & $\times$ & $\times$ & $\times$ & 41.8 & 21.2 & 38.9 & 57.3 \\
\midrule
\textbf{Ours} (Full Model) & $\checkmark$ & $\checkmark$ & $\checkmark$ & \textbf{46.2} & \textbf{33.0} & \textbf{58.3} & \textbf{75.0} \\
\hspace{2mm} w/o Two-Stage Framework & $\times$ & $\checkmark$ & $\checkmark$ & 40.5 & 22.1 & 47.5 & 63.9 \\
\hspace{2mm} w/o Synthesized Data & $\checkmark$ & $\times$ & $\checkmark$ & 44.9 & 26.7 & 49.9 & 67.6 \\
\hspace{2mm} w/o RL Training & $\checkmark$ & $\times$ & $\times$ & 45.4 & 28.3 & 40.4 & 65.4 \\
\bottomrule
\end{tabular}
}
\end{table*}

\subsection{Ablation Study}
\label{sec: ablation}

We conduct a systematic ablation study to deconstruct our method and quantify the independent contribution of each core component on UniDoc-7B, as summarized in Table \ref{tab:main_ablation}.\looseness=-1

\textbf{Effect of the Two-Stage Framework.}
Removing the two-stage decomposition forces uniformly low resolution across all pages, collapsing retrieval and reasoning into a single under-resolved pass. The \texttt{w/o Two-Stage Framework} ablation degrades performance across all benchmarks. MMLongBench-Doc accuracy drops from 33.0 to 22.1, and the model even underperforms the \texttt{Base Model} on DUDE. The marginal gain over the base model (22.1 vs.\ 21.2) confirms that RL provides some benefit at low resolution but cannot compensate for the perceptual information lost when fine-grained reasoning pages receive no resolution boost.

\textbf{Effect of the Synthesized Data Engine.}
The \texttt{w/o Synthesized Data} ablation results in a consistent performance drop across all four benchmarks: MMLongBench-Doc falls from 33.0 to 26.7 and SlideVQA from 58.3 to 49.9. Public document datasets lack the complex, multimodal, and holistic reasoning queries our engine generates; removing them narrows the training distribution, degrading generalization on benchmarks that require cross-page structural reasoning.

\textbf{Effect of RL Training.}
The \texttt{w/o RL Training} setup preserves the two-stage architecture but lacks the learned policy to coordinate retrieval and reasoning. The \texttt{w/o RL Training} model, which serves as a zero-shot pipeline, already provides a notable improvement over the \texttt{Base Model}, increasing MMLongBench-Doc accuracy by 7.1 points. However, our \texttt{Full Model} further improves upon this zero-shot baseline by 4.7 points on MMLongBench-Doc and 17.9 points on SlideVQA. The gap is largest on SlideVQA, where multi-slide navigation demands a learned retrieval policy rather than heuristic page selection.

\section{Analysis}

\subsection{Analysis of Training Strategy}
\label{sec: analysis}
Our ablation study (Section~\ref{sec: ablation}) shows that \texttt{Direct RL} yields the best end-to-end performance for a shared-parameter model. This raises a practical question for our two-stage architecture: is it more effective to rely on autonomous exploration (\texttt{Direct RL}), or to provide explicit supervision by jointly fine-tuning the shared model on synthesized annotations from both stages prior to RL (\texttt{Unified SFT+RL})? We address the following:
\begin{itemize}[nosep]
    \item \textbf{RQ1:} What is the performance trade-off between the \texttt{Direct RL} and \texttt{Unified SFT+RL} strategies, and what explains this trade-off?
    \item \textbf{RQ2:} Are the stage-level reasoning annotations
    intrinsically effective, despite the trade-off observed in
    the unified model?
\end{itemize}

\subsubsection{RQ1: Performance Trade-off between Training
Strategies}
\begin{figure}[t]
  \centering
  \begin{minipage}[c]{0.48\linewidth}
    \centering
  \subfloat[Retrieval accuracy]{\label{fig:unified_analysis_2}\includegraphics[width=0.46\columnwidth]{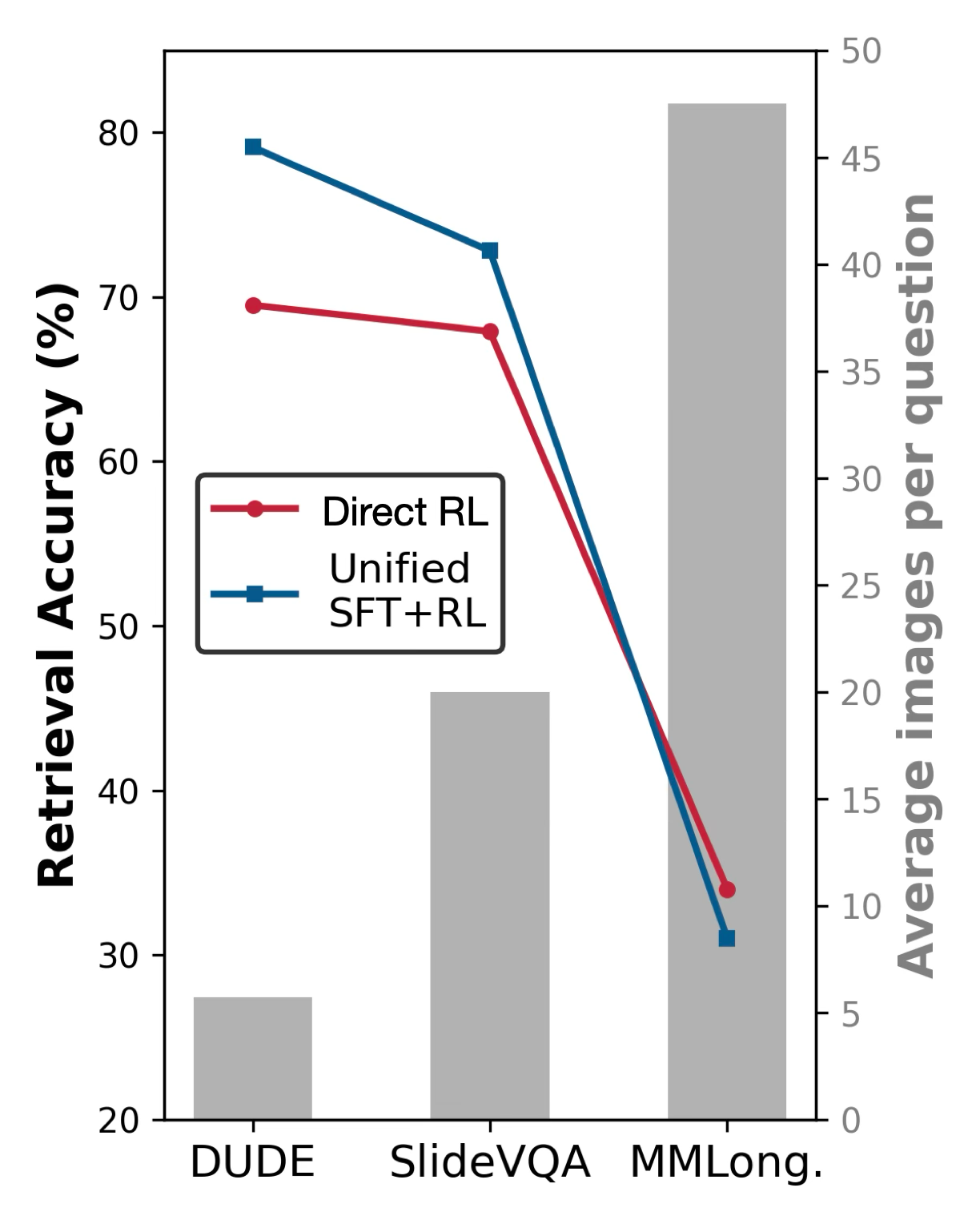}}
  \hspace{0.8em}
  \subfloat[Self-certainty and PPL]{\label{fig:unified_analysis_1}\includegraphics[width=0.46\columnwidth]{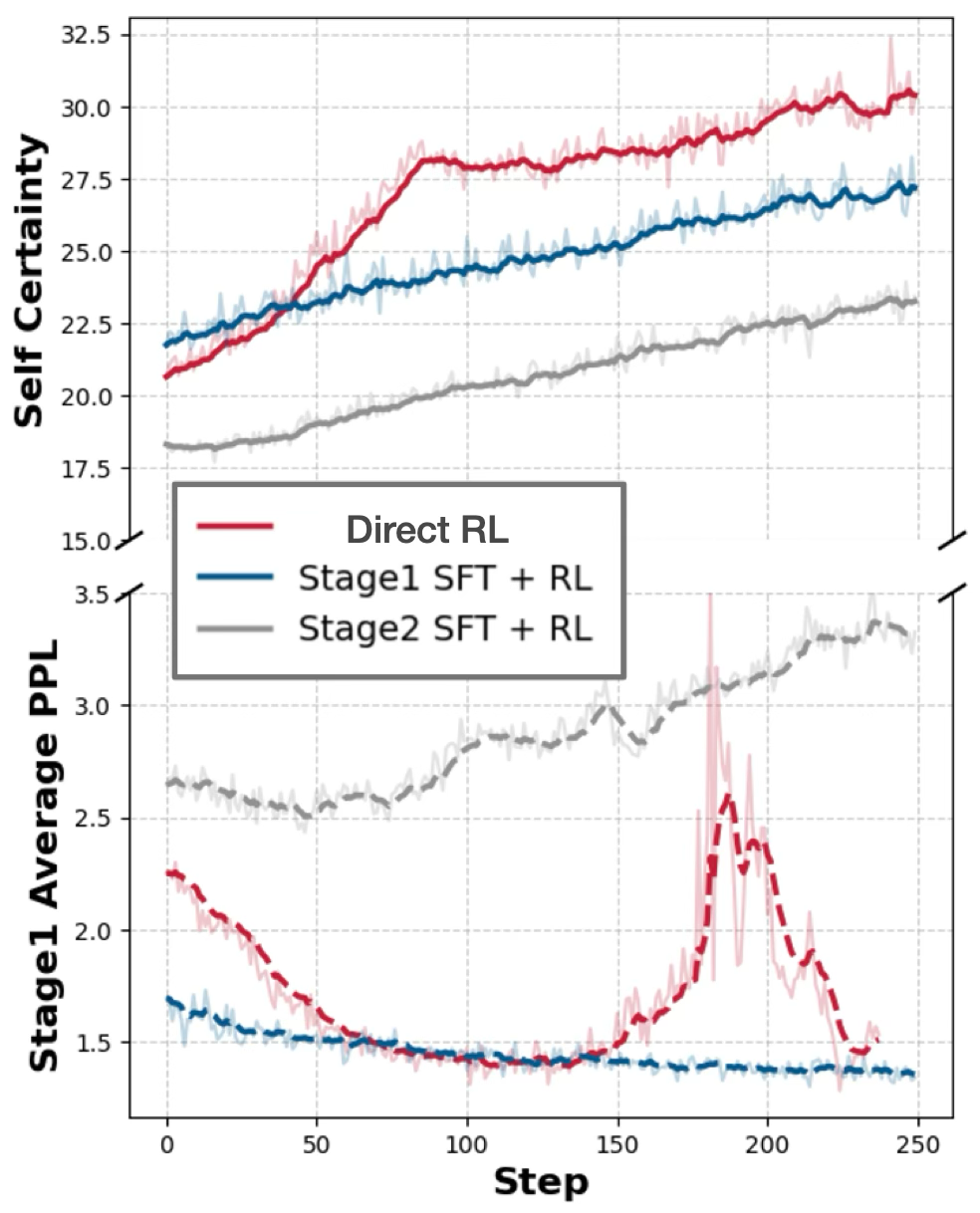}}
  \caption{\textbf{(a)} Retrieval Accuracy (Stage 1) across different context lengths. \textbf{(b)} Dynamics of three training recipes in RL training.}
  \label{fig:unified_analysis}
  \end{minipage}
  \hfill
    \begin{minipage}[c]{0.49\linewidth}
    \centering
    \subfloat[Stage 2 Performance]{\label{fig:stage2_barplot}\includegraphics[width=0.49\linewidth]{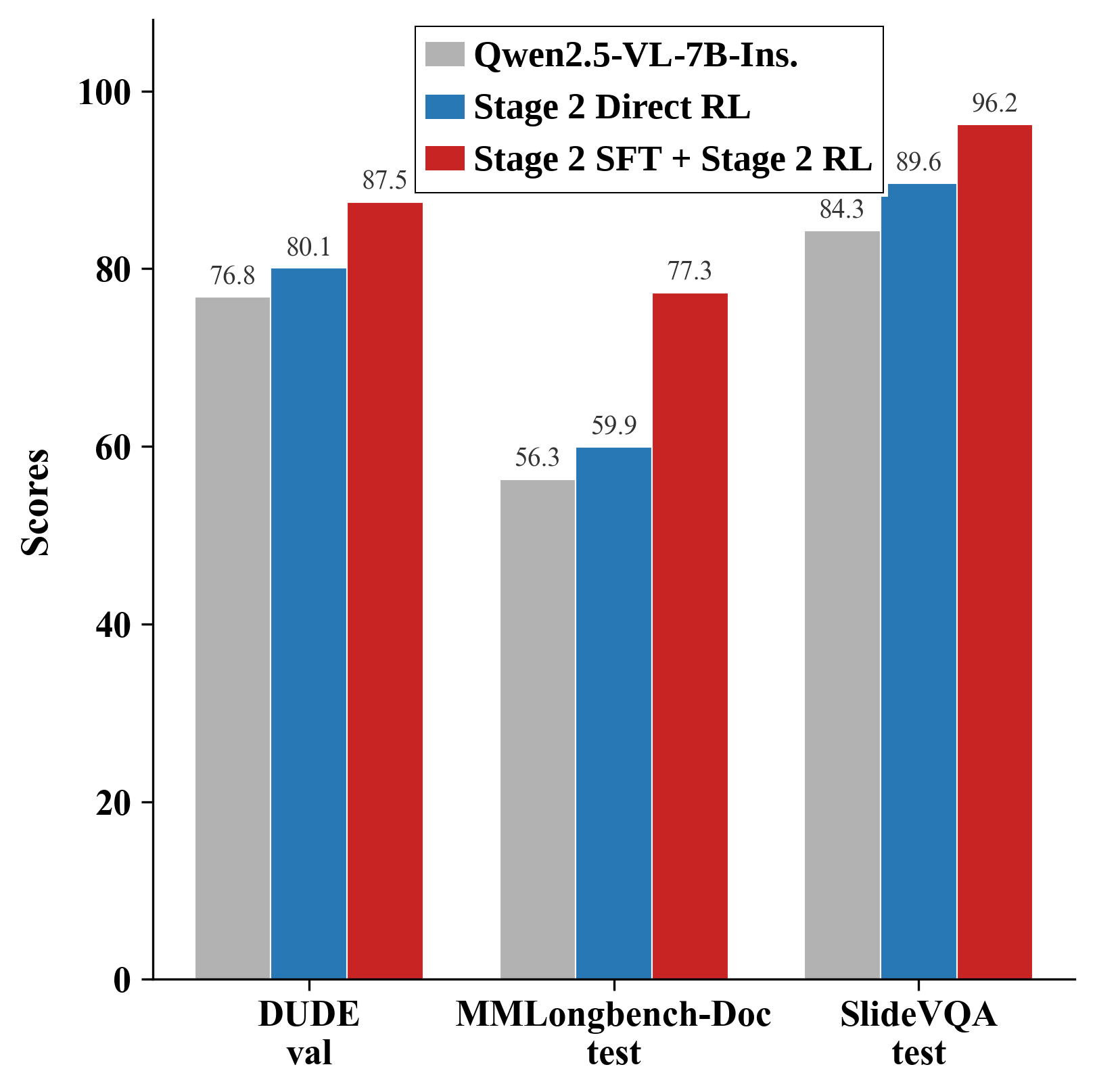}}
    \hspace{-0.2em}
    \subfloat[Length distribution]{\label{fig:length dist}\includegraphics[width=0.49\linewidth]{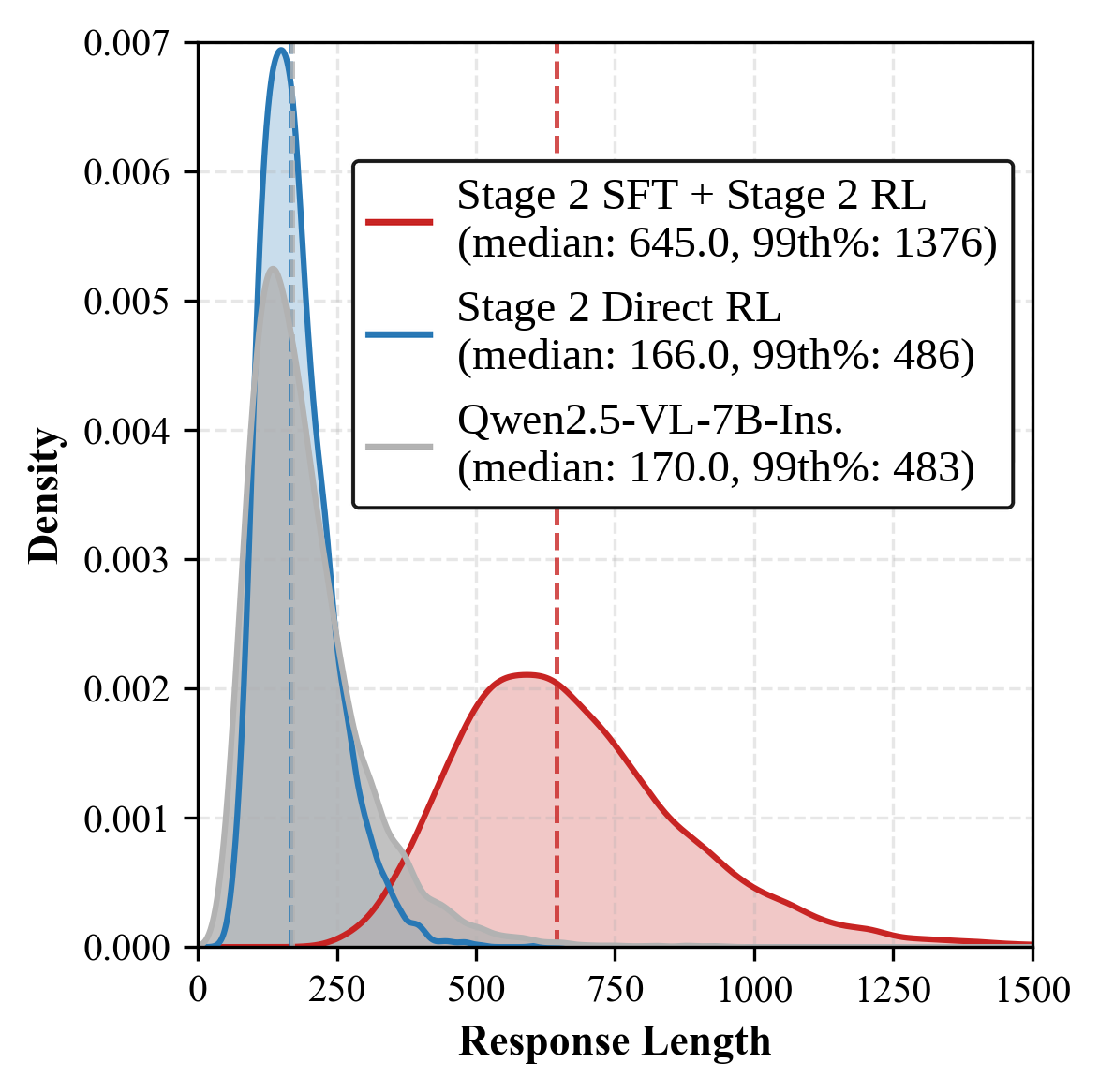}}
    \caption{\textbf{(a)} Performance comparison of different RL policies in a simulated Stage~2 setting. \textbf{(b)} Response length distributions for different training strategies in Stage~2.}
    \label{fig:stage2_analysis}
  \end{minipage}
\end{figure}

\textbf{The Unified SFT+RL policy exhibits a clear performance
dichotomy.} As shown in Table~\ref{tab:policy_tradeoff}, this
model achieves marginal gains on short-context reasoning tasks (e.g., DUDE, +1.0 ANLS) but suffers
significant degradation (-4.4 Acc.) on the long-context
MMLongBench-Doc. This weakness appears specific to long-context
processing. We confirm this by analyzing retrieval accuracy in
Figure~\ref{fig:unified_analysis_2}, which shows the
\texttt{Unified SFT+RL} model underperforms \texttt{Direct RL}
only on MMLongBench-Doc.

\begin{wraptable}{R}{0.50\linewidth}
  \vspace{-0.8\baselineskip}
  \centering
  \setlength{\tabcolsep}{5pt}
  \caption{Performance comparison of the final policy. The \texttt{Unified SFT+RL} exhibits a trade-off: marginal gains on short-context tasks but degradation on long-context MMLongBench-Doc.}
  \label{tab:policy_tradeoff}
  \resizebox{\linewidth}{!}{
    \begin{tabular}{lcccc}
      \toprule
      \textbf{Model} & \textbf{DUDE} & \textbf{MMLong.} & \textbf{SlideVQA} & \textbf{MP-DocVQA} \\
      & (ANLS) & (Acc.) & (F1) & (Acc.) \\
      \midrule
      \rowcolor{gray!25}
      \makecell[l]{Avg. Images \\ per Question} & 5.7 & 47.5 & 20.0 & 8.3 \\
      \midrule
      UniDoc-7B (Direct RL) & 46.2 & \textbf{33.0} & 67.9 & 75.0 \\
      Unified SFT+RL & \textbf{47.2} & 28.6 & \textbf{68.7} & \textbf{78.7} \\
      \midrule
      \textbf{Delta} & \textbf{+1.0} & \textbf{-4.4} & \textbf{+0.8} & \textbf{+3.7} \\
      \bottomrule
    \end{tabular}
  }
  \vspace{-0.8\baselineskip}
\end{wraptable}

\textbf{We hypothesize an internal
optimization tension.} The \texttt{Unified SFT} initialization forces
the model to jointly learn two competing capabilities: the
global, structure-level abstraction required by Stage~1 and the
local, pixel-level scrutiny required by Stage~2. These objectives pull shared parameters in opposing directions, producing an
imbalanced policy that subsequent RL optimization struggles
to reconcile.

\textbf{RL training dynamics are consistent with this hypothesis.} To isolate this tension, we examine models initialized with only single-stage SFT before RL. As shown in Figure~\ref{fig:unified_analysis_1}, the \texttt{Direct RL} policy, which learns both capabilities concurrently, maintains higher self-certainty than these single-stage models through training. The Stage~2 SFT objective appears detrimental to Stage~1: when evaluated on Stage~1 training data, the \texttt{Stage~2 SFT+RL} policy's perplexity (PPL) steadily increases during RL training, while the \texttt{Stage~1 SFT+RL} policy's PPL consistently decreases. SFT pre-training correlates with increasing perplexity of the Stage 2 SFT-initialized policy on Stage 1 data, consistent with conflicting optimization pressures between the two objectives. We hypothesize that Direct RL avoids this because it installs no such prior. It optimizes both stages from the base model under stage-aware rewards, with advantages normalized strictly within stage, so neither stage is fitted to a fixed target trajectory before the other is considered and the policy is never asked to reconcile an imbalanced initialization. We detail the computation of the self-certainty and perplexity metrics in Appendix~\ref{appd:training_metrics}.

\subsubsection{RQ2: Intrinsic Efficacy of Stage-Level Reasoning Annotations}

To isolate and validate the efficacy of the reasoning annotations, we trained two specialist models. Each specialist was trained and evaluated \textit{only} on its specific sub-task, using stage-specific proxy metrics.

\begin{figure}[t]
  \centering
  \begin{minipage}[t]{0.48\linewidth}
  \vspace{0pt}
    \centering
    \captionof{table}{Experimental results for different Stage 1 training strategies. Acc.\ (Accuracy) indicates if the retrieved set of page identifiers is an exact match, while Rec.\ (Recall) indicates if the retrieved set fully includes all relevant page numbers. FRAG~\citep{huang2025fragframeselectionaugmented} is the strongest RAG baseline listed in Table~\ref{tab:main_results}.}
    \label{tab:stage1}
    \small
    \setlength{\tabcolsep}{4pt}
    \renewcommand{\arraystretch}{1.0}
    \begin{tabular*}{\linewidth}{@{\extracolsep{\fill}}lcc@{}}
      \toprule
      \textbf{Methods} & Acc.$\uparrow$ & Rec.~$\uparrow$ \\
      \midrule
      \multicolumn{3}{@{}l}{\textit{MMLongBench-Doc (Stage 1)}} \\
      FRAG~\citep{huang2025fragframeselectionaugmented} & 3.70 & 54.81 \\
      Qwen2.5-VL-7B-Ins. & 37.78 & 42.96 \\
      \hspace{2mm} w/ stage1 Direct RL & 48.48 & 55.30 \\
      \hspace{2mm} w/ stage1 SFT & 53.33 & 63.48 \\
      \hspace{2mm} w/ stage1 SFT+RL & \textbf{56.30} & \textbf{65.93} \\
      \midrule
      \multicolumn{3}{@{}l}{\textit{SlideVQA (Stage 1)}} \\
      FRAG~\citep{huang2025fragframeselectionaugmented} & 4.97 & 71.92 \\
      Qwen2.5-VL-7B-Ins. & 59.73 & 62.21 \\
      \hspace{2mm} w/ stage1 Direct RL & 71.09 & 78.70 \\
      \hspace{2mm} w/ stage1 SFT & \textbf{76.67} & 78.84 \\
      \hspace{2mm} w/ stage1 SFT+RL & 75.41 & \textbf{80.47} \\
      \bottomrule
    \end{tabular*}
  \end{minipage}\hfill
  \begin{minipage}[t]{0.48\linewidth}
  \vspace{0pt}
  \centering
  \includegraphics[width=\linewidth]{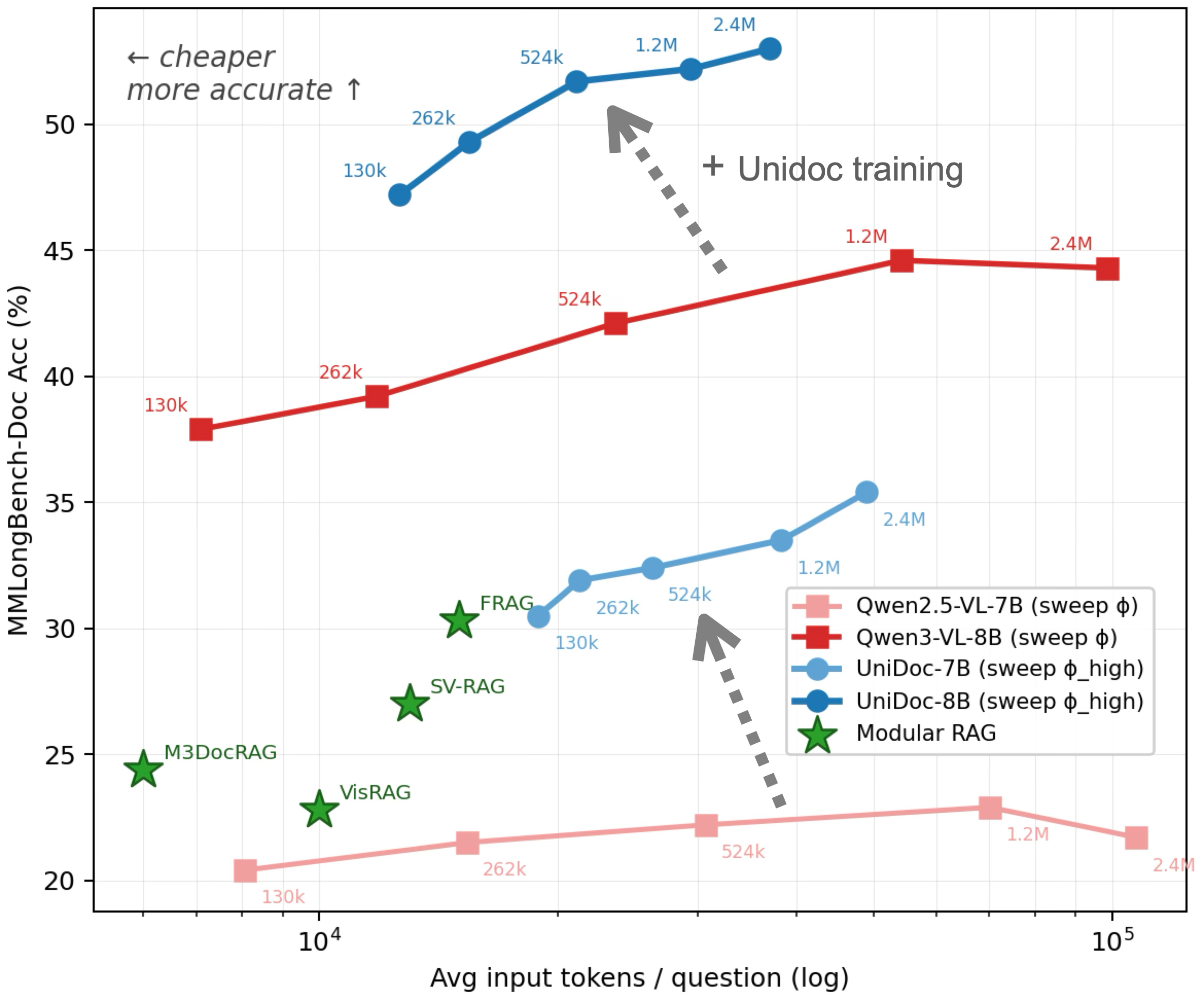}
  \caption{Accuracy--cost trade-off on MMLong\-Bench-Doc. The $x$-axis (log) is the average total input tokens per question, counting both UniDoc stages. UniDoc training shifts both backbones onto a better frontier.}
  \label{fig:pareto}
  \end{minipage}
\end{figure}

For Stage~1 (Table~\ref{tab:stage1}), the \texttt{Stage~1 SFT + Stage~1 RL} specialist outperforms other training recipes and strong RAG baselines in retrieval accuracy and recall, demonstrating that our reasoning annotations effectively enhance structural evidence localization. For Stage~2 (Figure~\ref{fig:stage2_barplot}), the \texttt{Stage~2 SFT + Stage~2 RL} specialist achieves an average +10.5 point gain over the \texttt{Stage~2 Direct RL} baseline. Additionally, the response length distribution in Figure~\ref{fig:length dist} shows that the model equipped with our adaptive layout reasoning pattern can calibrate its reasoning depth to different queries.

These results confirm that our synthesized data provides high-quality supervision for both stages individually, validating the effectiveness of our synthesis engine and the resulting dataset. The performance degradation observed in RQ1 is therefore not a failure of the data itself; we attribute it to optimization tension when jointly training both within a single model.

\subsection{Efficiency Analysis}
\label{sec:efficiency}
A central claim of UniDoc is that recovering the fidelity of full-resolution perception need not incur the token cost of monolithic high-resolution processing. We test this directly by measuring accuracy against realized input cost.

Figure~\ref{fig:pareto} plots MMLongBench-Doc accuracy against the average number of input tokens consumed to answer a question, counting the full cost of both stages for UniDoc, the low-resolution Fast Reading pass over all $N$ pages plus the high-resolution Focused Thinking pass over the localized pages. We trace each method's accuracy--cost curve by sweeping its perceptual budget: the single input resolution $\phi$ for monolithic VLMs, and the high-resolution budget $\phi_{\text{high}}$ for UniDoc. We focus on MMLongBench-Doc (avg.\ 47.5 pages) because token cost is load-bearing only in the long-context regime; on benchmarks with few pages per question the cost axis is not a meaningful constraint. We compare against the two monolithic backbones (Qwen2.5-VL-7B, Qwen3-VL-8B) and a set of modular RAG pipelines.

\textbf{UniDoc shifts the accuracy--cost frontier.} UniDoc strictly dominates the monolithic baselines: at matched or lower token cost, it attains substantially higher accuracy than the monolithic model it is built on. The effect is most pronounced on the stronger backbone, where the entire UniDoc-8B curve dominates the Qwen3-VL-8B curve, and UniDoc-8B's most economical configuration ($2.5\times10^4$ tokens, $\!47\%$) already exceeds the accuracy of Qwen3-VL-8B's most expensive configuration ($1.1\times10^5$ tokens,  $\!44\%$), i.e., achieving higher accuracy at roughly $4\times$ fewer input tokens.

\textbf{The gain comes from where resolution is spent, not how much.} Because Fast Reading reads all $N$ pages at low resolution and Focused Thinking applies high resolution only to the $k\!\ll\!N$ localized pages, raising $\phi_{\text{high}}$ improves accuracy while barely moving total token cost; the UniDoc curves are correspondingly compressed along the cost axis. A monolithic model, in contrast, pays any resolution increase across all $N$ pages. This also explains the baselines at the high end: both monolithic backbones saturate, and Qwen2.5-VL-7B degrades, as their budget grows—indiscriminately spending more tokens on full-document high-resolution processing yields diminishing or negative returns. The bottleneck is the read-everything paradigm, not the resolution ceiling.

\textbf{Comparison with RAG.} Modular RAG pipelines occupy a low-cost but performance-bounded regime: while inexpensive, their fragmented retrieval inherently caps accuracy. At a comparable order of token cost, UniDoc attains markedly higher accuracy; UniDoc-8B exceeds the strongest RAG baseline by roughly 20 points, without an external retriever or document chunking. Taken together, these results confirm that UniDoc's two-stage decomposition shifts cost from indiscriminate resolution to targeted retrieval, achieving higher accuracy at lower total token budget.

\subsection{Diagnostic Analysis}
\label{sec:diagnostic}

\begin{figure}[t]
    \centering
    \includegraphics[width=0.98\textwidth]{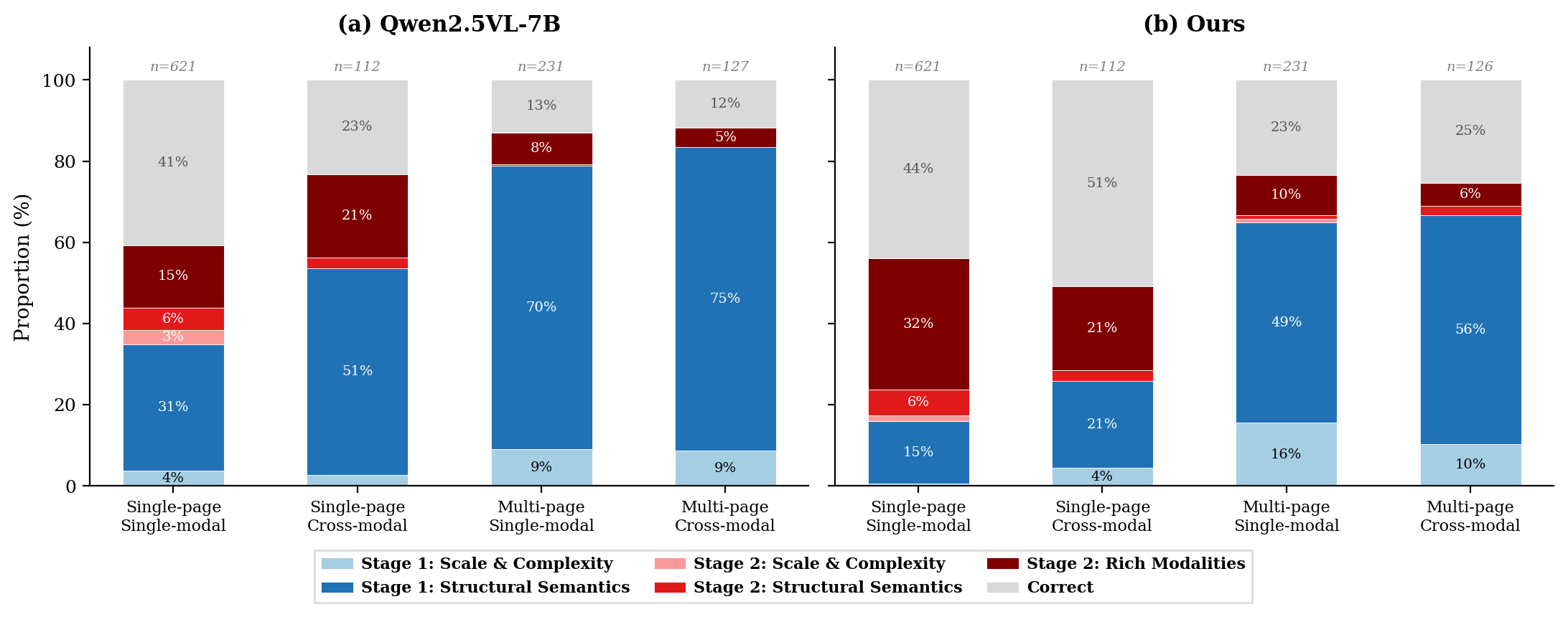}
    \caption{Error taxonomy on MMLongBench-Doc, comparing 
(a) the base model and (b) our trained model. Each error is 
attributed to a stage (Fast Reading or Focused Thinking) and a 
document characteristic.}
    \label{fig:error_decomposition}
\end{figure}

Our two-stage decomposition enables a natural error attribution mechanism: each failure can be traced to either Fast Reading or Focused Thinking, and further categorized by the underlying 
document characteristic (\textit{Rich Modalities}, \textit{Structural Semantics}, or \textit{Scale and Complexity}). Stage attribution is determined by whether the retrieved pages 
cover the ground-truth evidence set; characteristic attribution is obtained via automated analysis of the model's reasoning trace (details in Appendix~\ref{appd:error_taxonomy}). We apply this taxonomy to both the base model and our trained model on MMLongBench-Doc, grouping questions along two axes: page distribution (single-page vs.\ multi-page) and modality involvement (single-modal vs.\ cross-modal). Figure~\ref{fig:error_decomposition} compares the resulting error distributions.

Across both models, retrieval errors related to structural semantics consistently dominate, and this dominance intensifies with document complexity, confirming that structural understanding for evidence localization is a persistent bottleneck of document reasoning. Our training directly addresses this: retrieval errors related to structural semantics drop substantially across all categories (e.g., 75\% to 56\% in Multi-page Cross-modal, 51\% to 21\% in Single-page Cross-modal). Because Fast Reading and Focused Thinking operate sequentially, this improved retrieval shifts the error distribution in a predictable way. As more samples successfully reach the reasoning stage, fine-grained visual perception (Rich Modalities) is exposed as the new bottleneck (e.g., 15\% to 32\% in Single-page Single-modal).

These findings delineate two remaining dominant failure modes for document reasoning. First, the error shift toward \textit{Rich Modalities} failures suggests significant room for improvement in fine-grained visual perception. These errors occur even in single-modal queries, indicating fundamental perceptual failures (e.g., misreading table cell alignments or chart values) rather than just cross-modal integration deficits. This failure mode aligns naturally with recent advances in fine-grained visual tools for single-image understanding, yet their application to document-level reasoning remains largely unexplored.

Second, \textit{Structural Semantics} remains the dominant source of retrieval errors despite our training gains, far exceeding \textit{Scale and Complexity} in every category. We hypothesize this stems from a representational mismatch: documents are inherently hierarchical and cross-referential, yet models consume them as linear sequences of page images, discarding the structural topology. The path forward may lie not in longer contexts but in structured representations, whether through hierarchical memory that mirrors a document's organizational tree, explicit cross-reference graphs, or interactive navigation akin to GUI agents. Bridging this gap between linear consumption and structural understanding remains an important open challenge for future document reasoning.

\section{Conclusion}

UniDoc demonstrates that resolution asymmetry, the principle that evidence retrieval and multimodal reasoning demand fundamentally different perceptual granularities, can serve as a general decomposition strategy for long-document understanding. By separating low-resolution structural retrieval (Fast Reading) from high-resolution focused reasoning (Focused Thinking) within a single jointly trained model, UniDoc relaxes the scalability--fidelity tension without external retrieval modules or document fragmentation. Across two backbones, the framework shifts the accuracy--cost frontier: higher accuracy at up to $4\times$ fewer input tokens than monolithic processing.

Four limitations bound the current design. First, Fast Reading and Focused Thinking share parameters but exchange no cross-stage gradients, leaving end-to-end optimization to RL reward alone. Second, the analysis in RQ1 suggests an optimization tension between stage-level SFT and joint RL, limiting how much explicit supervision a single parameter set can leverage. Third, retrieval operates at page-level granularity, which cannot localize sub-page evidence such as a single table row or figure panel. Fourth, Focused Thinking cannot recover a page omitted by Fast Reading; retaining non-selected pages at low resolution mitigates this but lowers accuracy when retrieval is already complete.

Beyond these limitations, our error analysis points toward two open problems. Structured document representations, such as hierarchical memory, cross-reference graphs, or interactive navigation, could replace the linear page-sequence assumption that dominates current retrieval errors. Separately, integrating fine-grained visual tools into the Focused Thinking stage could address the Rich Modalities failures that emerge once retrieval accuracy improves. Both directions extend the resolution-asymmetry principle from a two-stage pipeline toward a richer hierarchy of perceptual operations.

\section*{Broader Impact Statement}

UniDoc lowers the cost of automated question answering over long documents. The same capability that supports legitimate uses such as auditing and legal review also makes large-scale analysis of sensitive documents cheaper, including personal, medical, and proprietary material that would otherwise require human effort to process. Reducing the cost of extracting information from documents at scale can therefore enable surveillance or unauthorized analysis of private records. Our training data is synthesized from publicly available academic papers and financial filings, and models trained on it may reflect biases present in those sources and in the vision-language model used for synthesis.

\bibliography{main}
\bibliographystyle{tmlr}

\appendix
\clearpage

\appendix

\section{Benchmarks}
\label{appd:benchmarks}
Our evaluation consists of 4 challenging document understanding benchmarks, selected to test diverse reasoning capabilities. All 4 benchmarks are multi-page, mimicking the real-life situation.

\begin{itemize}
    \item \textbf{MMLongBench-Doc~\citep{ma2024mmlongbenchdocbenchmarkinglongcontextdocument}:} A challenging benchmark designed to evaluate MLLMs on long-form, multi-modal documents. It consists of documents from diverse sources, including research papers, brochures, guidebooks, and financial reports. With an average document length of 47.5 pages, it specifically tests a model's ability to perform long-context, compositional reasoning.

    \item \textbf{DUDE~\citep{vanlandeghem2023documentunderstandingdatasetevaluation}:} A comprehensive benchmark for Document Understanding Dataset and Evaluation. It aggregates tasks from 12 real-world document datasets, focusing on a model's ability to perform fine-grained information extraction from visually-rich documents like invoices, receipts, and forms.
    
    \item \textbf{SlideVQA~\citep{tanaka2023slidevqadatasetdocumentvisual}:} A benchmark for visual question answering on presentation slides. This dataset specifically challenges a model's ability to understand complex layouts, parse charts and diagrams, and reason about the interplay between textual and visual elements in a slide deck.

    \item  \textbf{MP-DocVQA~\citep{tito2023hierarchicalmultimodaltransformersmultipage}:} A benchmark for visual question answering on multi-page documents. This dataset specifically challenges a model's ability to perform long-context reasoning, navigate across an entire scanned document to find the relevant information, and accurately locate both the answer text and the specific page number where it was found.
\end{itemize}

Standard evaluation scripts often penalize correct but verbose answers (e.g., mismatches due to punctuation). To address this, we extend the LLM-based answer extraction pipeline used in MMLongBench-Doc~\citep{ma2024mmlongbenchdocbenchmarkinglongcontextdocument} to all evaluated benchmarks. 

Before computing the metrics, we employ an extractor (e.g., GPT-4o) to parse the predicted answer from the raw model response, ensuring it aligns with the format of the ground truth. We then utilize the official scripts for final metric calculation. Consequently, for DUDE~\citep{vanlandeghem2023documentunderstandingdatasetevaluation}, we report the ANLS metric; for MP-DocVQA~\citep{tito2023hierarchicalmultimodaltransformersmultipage}, we report accuracy; for SlideVQA~\citep{tanaka2023slidevqadatasetdocumentvisual}, we report both accuracy and F1 score; for MMLongBench-Doc, we follow the official guideline, using GPT-4o as the answer extractor and report both accuracy and F1 score.

Following standard practice in document understanding evaluation, we report single-run results; with N > 1,000 per benchmark, the standard error of accuracy is below 0.5 points.

\begin{figure}[t]
  \centering
  \includegraphics[width=0.75\linewidth]{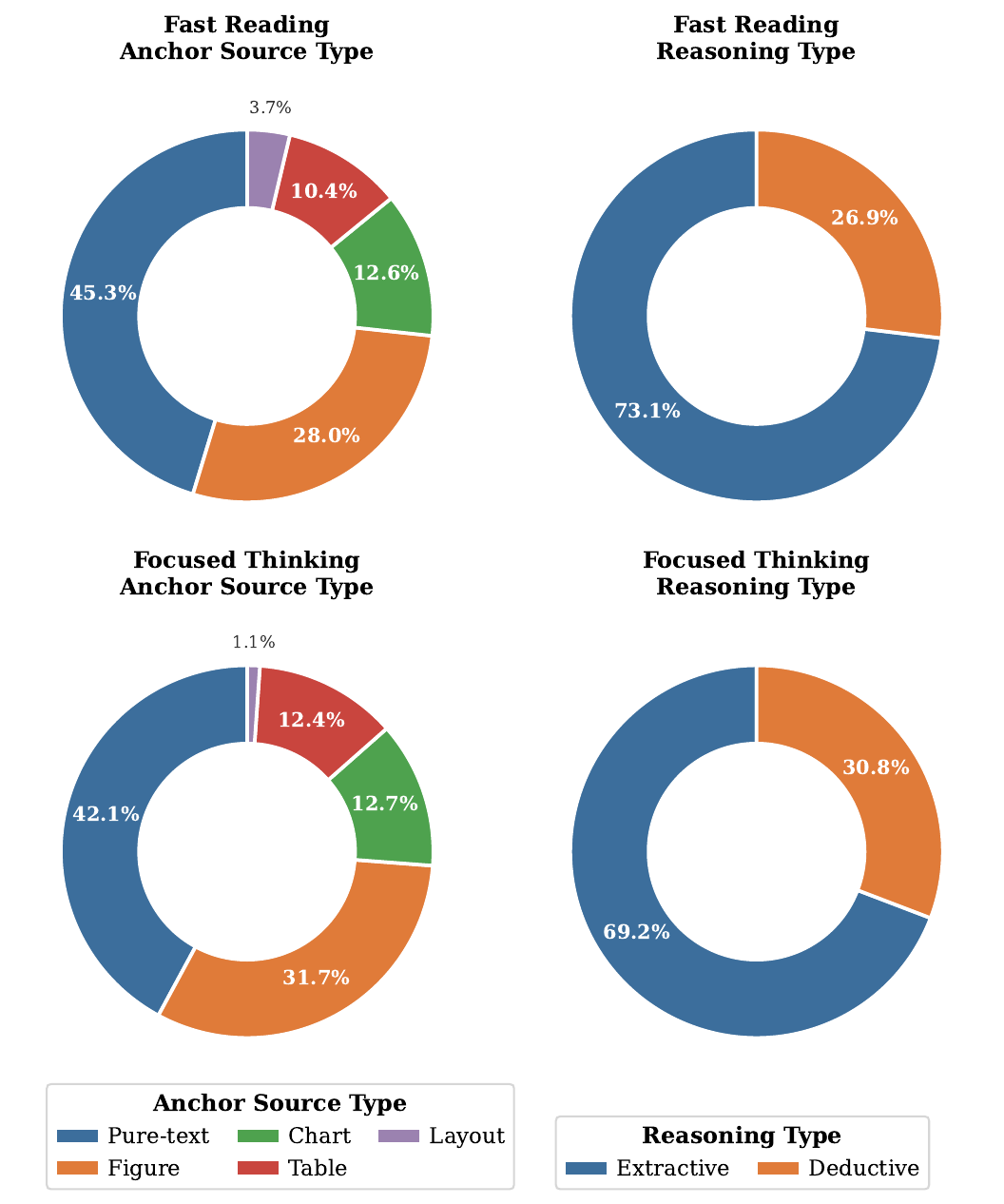}
  \captionof{figure}{Data Engine query distribution.}
  \label{fig: different categories}
\end{figure}

\section{Data Engine Details}
\label{appd: query syn}

\subsection{Source Corpus}
\label{app:corpus}

Our synthesis engine operates on two document collections we assembled: 10K papers crawled from arXiv and 10K financial filings crawled from SEC EDGAR. The two are complementary in the structures they exercise. arXiv papers carry dense cross-references between narrative text and figures, tables, and equations, which is what causes the evidence for a single query to distribute across distant pages. SEC filings are longer and more table-oriented, with numerical evidence embedded in dense financial statements and interpreted in surrounding disclosure text. Together they cover both reasoning demands the framework targets: structural navigation across a document and fine-grained multimodal interpretation within a page. Each ground-truth evidence set is determined by the pages on which the anchor and its referencing passages actually appear, so the stage-level annotations are grounded in document structure rather than assigned post hoc.

\subsection{Anchor Types and Processing Strategies}
Our data synthesis engine, as outlined in Algorithm \ref{alg:synthesis}, operates on the principle of multi-modal anchors, such as figures in scientific papers or tables in financial reports. In this section, we first present the statistical distribution of these anchor types and subsequently detail the specific processing strategies employed for each type within our synthesis pipeline.

As illustrated in Figure \ref{fig: different categories}, our multi-modal anchors encompass diverse categories, including figures, charts, and tables. Notably, we also incorporate \textit{pure-text} regions as a distinct anchor type, which constitutes a significant proportion of the dataset. This design choice is motivated by the fact that a paragraph's semantic meaning is often intrinsically linked to its specific spatial layout within the document, thereby functioning as a quasi-visual element.

\textbf{Pure-Text.} For pure-text anchors, we supply the generator with both the OCR transcription of the text and the full document page image containing the text. The inclusion of the full page view is critical, as the synthesized query may necessitate reasoning based on the specific layout information or spatial context of the target paragraph.

\textbf{Figures and Charts.} For visual anchors such as figures and charts, we extract the specific visual crop of the anchor. Crucially, we also retrieve and include associated textual context that references these visual elements, even if such text is spatially distributed across different pages, to ensure the generated query is grounded in complete information.

\textbf{Tables.} For tabular anchors, simply providing the visual crop is often insufficient due to the density of information. Therefore, in addition to the cropped table image and its referencing text, we explicitly provide the OCR parsing result of the table. This supplementary textual representation significantly mitigates the difficulty of structural interpretation for the generator, leading to higher-quality query synthesis.

\textbf{Layout.} Layout anchors target cases where evidence lies in the visual structure of the page itself, such as heading hierarchies, font weight differences, and spatial grouping. We provide the generator with the full page image alongside an explicit description of these structural cues, ensuring the synthesized query requires reasoning about document organization 
rather than content alone.

Together, these four anchor types ensure that our synthesized dataset covers a broad spectrum of document elements and reasoning demands, from pure-text comprehension to cross-modal synthesis and structural layout understanding.

\subsection{Query Reasoning Types}
While the diversity of anchor types demonstrates the breadth of our data engine, the depth of reasoning required by the synthesized queries is equally critical. We categorize the synthesized queries into two distinct reasoning paradigms: \textit{extractive} and \textit{deductive}.

\textbf{Extractive Queries.} These queries do not necessarily require complex post-processing. Once the model successfully performs standard grounding and extraction of the relevant content, the answer is typically obtained directly.

\textbf{Deductive Queries.} Conversely, deductive queries require the model to perform further reasoning \textit{after} extracting the information. This primarily involves arithmetic operations (e.g., calculating the difference between two numerical values) or logical inference processes (e.g., identifying the maximum value within a specific table column).

We illustrate the statistical distribution of these reasoning types in Figure \ref{fig: different categories}. Extractive queries, which constitute the majority of existing datasets, account for approximately 70\% of our corpus, while the more challenging deductive queries comprise the remaining 30\%.

\section{Design Rationale for Stage-Level Reasoning Annotations}
\label{appd: pattern_design}

This appendix provides the detailed methodological justifications for the ``Reasoning Annotation'' designs summarized in Section \ref{sec:thinking_patterns}.

\subsection{Stage 1: Rationale for Context-Aware Localization}
\label{appd:stage1}

\paragraph{Design Philosophy}
We posit that ``Fast Reading'' (Stage 1) is fundamentally a \textbf{reasoning-centric localization task}, not a passive filtering task. Traditional retrieval methods (i.e., RAG), which are limited to calculating local query-content similarity, are structurally incapable of comprehending a document's holistic structure. To transcend this ``context-blind'' limitation, the model must learn to be context-aware. A document's underlying logical flow—the very ``evidence'' required for localization—is often not a single piece of text but rather the \textit{logical interplay} of its heterogeneous components (e.g., text, tables, figures). Therefore, to successfully find the correct context, the model must first learn to think holistically about these document-wide connections.

\paragraph{Synthesis Process}
To synthesize this ``context-aware'' SFT data ($\tau_1$), our pipeline is designed to generate explicit reasoning trajectories that derive the set of relevant page identifiers $\mathcal{P}_{\text{gt}}$. 

Given a query $q$ and a document $d$, we construct $\tau_1$ via a \textbf{grounded rationalization process}. We first partition the full document $d$ into a sequence of chunks $\{c_i\}_{i=1}^n$, ensuring that the concatenation of these chunks reconstructs the original document. We then iteratively present each chunk $c_i$ to a powerful VLM, providing \textit{both} the full multimodal content of the chunk and the subset of ground-truth identifiers $\mathcal{P}_{\text{gt}, i}$ contained therein. The VLM is prompted to perform a context-aware rationalization: it must reverse-engineer the strategic steps and logical connections ($\rho_i$) that justify \textit{why} $\mathcal{P}_{\text{gt}, i}$ are the correct pages. Crucially, the model is instructed to base its reasoning explicitly on the logical flow and the interplay of textual, visual, and structural evidence within $c_i$. These high-fidelity partial thoughts $\{\rho_i\}$ are subsequently aggregated. A final consolidation prompt is applied to synthesize them into a single, coherent holistic reasoning trajectory $\tau_1$. This final step explicitly encourages the model to uncover and articulate relationships \textit{across} the chunks. Thus, while the synthesis of each $\rho_i$ ensures intra-chunk local coherence, the consolidated $\tau_1$ enforces global, cross-chunk context awareness.
\nocite{wang2026badseeingbadthinking,searchgen,rationalrewards}

\paragraph{The Trade-off of Chunking Granularity} A critical hyperparameter in this pipeline is the chunking granularity, determined by the number of chunks $n$. This parameter governs the fundamental trade-off between the \textbf{continuity of context-awareness} and the \textbf{perceptual resolution} of the content. When $n$ is large (implying smaller chunks with fewer pages), the VLM can process each chunk at a higher effective resolution, capturing finer visual details. However, excessive chunking leads to severe context fragmentation. As the chunk size decreases (i.e., $n$ approaches the total page count), the framework risks collapsing into a fragmented, RAG-like paradigm governed by local similarity, thereby losing the holistic structural understanding. Conversely, a small $n$ preserves global context but limits the resolution available for each page. To balance this trade-off, our synthesis pipeline standardizes the chunk size to 5 pages. Consequently, the number of chunks is dynamically set as $n = \lceil \text{len}(d) / 5 \rceil$. This configuration ensures sufficient visual resolution for analyzing layout and figures while maintaining enough continuous context to preserve structural semantics.

\subsection{Stage 2: Rationale for Adaptive Grounding-Reasoning}
\label{appd:stage2}

\paragraph{Design Philosophy}
For the ``Focused Thinking'' stage, we characterize the task as a sequential process of localization and extraction: the agent must iteratively decide \textit{where} to look, extract the relevant information, and then determine the next location to attend to.

In this loop, the critical bottleneck is not extraction; provided the model is correctly grounded, extracting text from a specific location is relatively trivial. The true challenge, and the primary locus of reasoning, lies in the \textbf{grounding step itself}—the strategic decision-making process of determining where to look based on the query and the information gathered thus far.

We posit that this grounding capability cannot be monolithic; it must be \textbf{adaptive}. The required grounding pattern is intrinsically correlated with the difficulty of the query and the visual complexity of the document. We distinguish between simple queries, which demand a single turn of grounding and extraction, and complex queries, which necessitate multi-step, compositional paths. Training a model on a static reasoning pattern is suboptimal: a simple pattern fails on complex problems, while a complex pattern is computationally inefficient and prone to ``overthinking'' on simple ones.

\paragraph{Synthesis Process}
To instill this adaptive prior, our SFT dataset ($\tau_2$) is constructed to teach the model to associate query complexity with the appropriate grounding pattern. We achieve this through a two-step ``categorize-then-generate'' pipeline.

First, we \textbf{categorize} our synthesized queries (from Section \ref{sec:data synthesis}) based on their intrinsic difficulty relative to our base model (Qwen2.5-VL-7B-Instruct). We operationalize difficulty by testing solvability under progressively richer prompting strategies:
\begin{itemize}
    \item \textbf{Level 1 (Easy):} Queries correctly answered by the model using a simple, direct prompt.
    \item \textbf{Level 2 (Medium):} Queries that fail with direct prompting but are correctly solved when the model is instructed with a detailed Chain-of-Thought (CoT) prompt.
    \item \textbf{Level 3 (Hard):} Queries that fail even with advanced CoT prompting.
\end{itemize}

Second, we \textbf{generate} a distinct SFT target pattern ($\tau_2$) tailored to each difficulty level:
\begin{itemize}
    \item \textbf{For Level 1 (Easy):} The SFT target is the direct response generated by the model itself. This implicitly teaches a concise ``direct-ground-and-extract'' pattern, encouraging efficiency for simple tasks.
    \item \textbf{For Level 2 (Medium):} The SFT target is the successful step-by-step CoT trace obtained during the categorization phase. This explicitly models a linear ``ground-then-see-then-ground-again'' sequence required for multi-hop reasoning.
    \item \textbf{For Level 3 (Hard):} For these most complex queries, we adopt the REER methodology~\citep{wang2025reverseengineeredreasoningopenendedgeneration}. We employ an iterative, error-correcting refinement process: (1) we prompt a powerful VLM to generate an initial reasoning trace; (2) a ``verifier'' prompt critiques the trace to identify logical flaws or missing grounding steps; and (3) the VLM refines the trace based on this critique. This cycle repeats until a verifiable, complex grounding-reasoning trajectory is produced.
\end{itemize}

\section{Training Details}

\subsection{Supervised Fine-Tuning}

\paragraph{Hyperparameters.}
We employ Llama-Factory~\citep{zheng2024llamafactory} as the LLM training platform. Table \ref{tab:hp_stage1}, Table \ref{tab:hp_stage2} and Table \ref{tab:hp_mix} show our training hyperparameters in supervised fine-tuning. We use 8 A100 80G GPUs for training.

\begin{table}[!h]
    \centering
    \caption{Hyperparameters for supervised fine-tuning in \textbf{Stage~1}.}
    \begin{tabular}{ll}
        \toprule
        Parameter        & Value                                           \\
        \midrule
        Per device train batch size & 1                                              \\
        Gradient accumulation steps & 4                                              \\
        Learning rate    & 1.0e-5                                          \\
        Number of epochs & 5.0                                             \\
        LR scheduler     & cosine                                          \\
        Warmup ratio     & 0.1                                             \\
        Precision        & bf16                                            \\
        Image max pixels        & 262144                                           \\
        Freeze vision tower        & true                                          \\
        Freeze multi-modal projector        & true                                           \\
        Freeze language model       & false                                          \\
        \bottomrule
    \end{tabular}
    \label{tab:hp_stage1}
\end{table}

\begin{table}[!h]
    \centering
    \caption{Hyperparameters for supervised fine-tuning in \textbf{Stage~2}.}
    \begin{tabular}{ll}
        \toprule
        Parameter        & Value                                           \\
        \midrule
        Per device train batch size & 2                                              \\
        Gradient accumulation steps & 4                                              \\
        Learning rate    & 2.0e-6                                          \\
        Number of epochs & 2.0                                             \\
        LR scheduler     & cosine                                          \\
        Warmup ratio     & 0.1                                             \\
        Precision        & bf16                                            \\
        Image max pixels        & 1204224                                           \\
        Freeze vision tower        & true                                          \\
        Freeze multi-modal projector        & true                                           \\
        Freeze language model       & false                                          \\
        \bottomrule
    \end{tabular}
    \label{tab:hp_stage2}
\end{table}

\begin{table}[!h]
    \centering
    \caption{Hyperparameters for supervised fine-tuning in mixed training.}
    \begin{tabular}{ll}
        \toprule
        Parameter        & Value                                           \\
        \midrule
        Per device train batch size & 1                                              \\
        Gradient accumulation steps & 8                                              \\
        Learning rate    & 1.0e-5                                          \\
        Number of epochs & 2.0                                             \\
        LR scheduler     & cosine                                          \\
        Warmup ratio     & 0.1                                             \\
        Precision        & bf16                                            \\
        Image max pixels        & Adaptive                                           \\
        Freeze vision tower        & true                                          \\
        Freeze multi-modal projector        & true                                           \\
        Freeze language model       & false                                          \\
        \bottomrule
    \end{tabular}
    \label{tab:hp_mix}
\end{table}

\paragraph{Training Data.}
For Stage~1, we adopt the same query instances from our synthesized data (Section~\ref{sec:data synthesis}), DUDE~\citep{vanlandeghem2023documentunderstandingdatasetevaluation} and SlideVQA~\citep{tanaka2023slidevqadatasetdocumentvisual}. Then, we synthesize thinking trajectories for these queries following the Fast Reading reasoning annotations detailed in Section~\ref{sec:thinking_patterns}.
For Stage~2, we adopt queries from our data engine and SlideVQA~\citep{tanaka2023slidevqadatasetdocumentvisual}. Then, we synthesize thinking trajectories for these queries following the adaptive-grounding reasoning annotations detailed in Section~\ref{sec:thinking_patterns}.
The distribution of query sources in these two stages is detailed in Table~\ref{tab:query source}.

\begin{table}[!h]
    \centering
    \caption{Query Sources in SFT Training}
    \begin{tabular}{l l r} 
        \toprule
        Stage   & Source                         & Nums  \\
        \midrule
        Stage 1 & DUDE                           & 5148  \\
        Stage 1 & SlideVQA                       & 3722  \\
        Stage 1 & Synthesized from Data Engine   & 2580  \\
        Stage 2 & SlideVQA                       & 13414 \\
        Stage 2 & Synthesized from Data Engine   & 4321  \\
        \bottomrule
    \end{tabular}
    \label{tab:query source}
\end{table}

\subsection{Reinforcement Learning}
\label{appd: reward design}

\paragraph{Hyperparameters.} We employ veRL~\citep{sheng2024hybridflow} as the training platform. Table \ref{tab:hp_rl} shows our training hyperparameters in the RL stage. All RL training runs are performed on 8 A800 GPUs or 8 H200 GPUs.

\begin{table}[!h]
    \centering
    \caption{Hyperparameters for reinforcement learning.}
    \begin{tabular}{ll}
        \toprule
        Parameter        & Value                                           \\
        \midrule
        Train batch size & 64                                              \\
        mini batch size    & 32                                          \\
        n resp per prompt & 8                                             \\
        max prompt length     & 14500                                          \\
        max response length     & 2500                                             \\
        actor learning rate      & 1.0e-6                                           \\
        advantage estimator      & GRPO                                        \\
        use kl in reward        & False                                        \\
        use kl loss             & True                                        \\
        kl loss coefficient      & 0.001                                        \\
        repetition penalty       & 1.00 \\
        temperature              & 1.0  \\
        use dynamic bsz          & True  \\
        max num gen batches      & 3 \\
        \bottomrule
    \end{tabular}
    \label{tab:hp_rl}
\end{table}

\paragraph{Training Data.}
For stage 1, we use the dataset in SFT training from our data synthesis engine. For stage 2,
there are two sources of our training queries, queries synthesized from our data engine and queries from DUDE training set. Specifically, we use 20k queries synthesized from our data engine and 5.7k queries from DUDE training dataset.

\paragraph{Reward Design.}
We define a stage-aware reward function tailored to the distinct 
output formats of each stage.

For Fast Reading, the model outputs a predicted page set 
$\mathcal{P}_{\text{pred}}$ evaluated against the ground-truth 
set $\mathcal{P}_{\text{gt}}$ via two binary metrics:
\begin{itemize}[nosep]
    \item \textbf{Accuracy:} Exact set match, 
    $\text{Acc} = \mathbb{I}(\mathcal{P}_{\text{pred}} = 
    \mathcal{P}_{\text{gt}})$.
    \item \textbf{Recall:} Ground-truth coverage, 
    $\text{Rec} = \mathbb{I}(\mathcal{P}_{\text{gt}} \subseteq 
    \mathcal{P}_{\text{pred}})$.
\end{itemize}
The Fast Reading reward averages both metrics, rewarding precision 
and completeness jointly:
$$
R_{\text{FR}} = \tfrac{1}{2}\bigl(\text{Acc}(\mathcal{P}_{\text{pred}}, 
\mathcal{P}_{\text{gt}}) + \text{Rec}(\mathcal{P}_{\text{pred}}, 
\mathcal{P}_{\text{gt}})\bigr)
$$

For Focused Thinking, the model produces a free-form answer 
$\hat{y}$ that may have diverse valid verbalizations, making 
simple string matching insufficient. We therefore adopt an 
LLM-as-judge that scores the semantic correctness and factual 
alignment of $\hat{y}$ against the ground-truth answer $y$:
$$
R_{\text{FT}} = \text{LLM-Judge}(q,\; \hat{y},\; y) \in [0, 1]
$$
The judge prompt is detailed in Table~\ref{appd: llmjudge prompt}.

\paragraph{Optimization.}
We optimize the unified policy $\pi_\theta$ via reinforcement 
learning on a mixed-stage dataset 
$\mathcal{D}_{\text{RL}} = \mathcal{D}_{\text{FR}} \cup 
\mathcal{D}_{\text{FT}}$, where $\mathcal{D}_{\text{FR}}$ and 
$\mathcal{D}_{\text{FT}}$ contain queries requiring Fast Reading 
and Focused Thinking trajectories, respectively. Both task types 
are sampled jointly within each batch.

Following the GRPO framework~\citep{shao2024deepseekmathpushinglimitsmathematical}, 
for each query $q \sim \mathcal{D}_{\text{RL}}$ we sample a group 
of $G$ trajectories $\{\tau_i\}_{i=1}^G$ and compute a normalized 
advantage for each:
\begin{equation*}
  \hat{A}_i = \frac{R(\tau_i) - \mu_G}{\sigma_G}, \quad
  \mu_G = \frac{1}{G}\sum_{j=1}^{G} R(\tau_j), \quad
  \sigma_G = \sqrt{\frac{1}{G}\sum_{j=1}^{G}
  \left(R(\tau_j) - \mu_G\right)^2}
\end{equation*}
The policy is optimized by maximizing the clipped surrogate 
objective with KL regularization:
\begin{equation*}
  \max_{\pi_\theta}\;
  \mathbb{E}_{\substack{q \sim \mathcal{D}_{\text{RL}},\\
  \tau \sim \pi_{\theta_{\text{old}}}(\cdot|q)}}
  \left[\,
    \min\!\Big(
      r_\theta(\tau)\,\hat{A}_i,\;\;
      \operatorname{clip}\!\big(r_\theta(\tau),\,
      1{-}\epsilon,\, 1{+}\epsilon\big)\,\hat{A}_i
    \Big)
    - \beta\, D_{\text{KL}}\!\big(
      \pi_\theta(\cdot|q)
      \,\|\,
      \pi_{\text{ref}}(\cdot|q)
    \big)
  \,\right]
\end{equation*}
where $r_\theta(\tau) = \pi_\theta(\tau|q) / 
\pi_{\theta_{\text{old}}}(\tau|q)$ is the importance sampling 
ratio, $\pi_{\text{ref}}$ is the reference policy, and $\beta$ 
controls KL regularization strength.

\subsection{Error Taxonomy Methodology}
\label{appd:error_taxonomy}

We propose a two-dimensional error taxonomy that classifies each 
incorrect prediction along two axes: \textbf{error stage} (whether 
the failure occurs in Fast Reading or Focused Thinking) and 
\textbf{error characteristic} (which document characteristic 
causes the failure). Below we detail each step.

\paragraph{Step 1: Error Stage Attribution.}
For each incorrect prediction, we determine whether the error 
originates from \textit{Fast Reading} (retrieval failure) or 
\textit{Focused Thinking} (reasoning failure) using a 
deterministic heuristic based on page-level retrieval coverage. 
Let $\mathcal{P}_{\text{gt}}$ denote the set of ground-truth 
evidence pages and $\mathcal{P}_{\text{pred}}$ denote the set of 
pages retrieved by the model. The stage is assigned as:
$$
\text{Error Stage} =
\begin{cases}
\text{Focused Thinking} & \text{if } \mathcal{P}_{\text{gt}} 
\subseteq \mathcal{P}_{\text{pred}} \\
\text{Fast Reading} & \text{otherwise}
\end{cases}
$$
If the model retrieved all ground-truth evidence pages (possibly 
among others), the retrieval stage succeeded and the error must 
have occurred during reasoning. Conversely, if any ground-truth 
page is missing, we attribute the error to Fast Reading.

\paragraph{Step 2: Error Characteristic Attribution.}
We classify each error into exactly one of the three document 
characteristics identified in our framework: \textit{Rich 
Modalities}, \textit{Structural Semantics}, and \textit{Scale 
and Complexity}.

\begin{itemize}[nosep]
    \item \textbf{Rich Modalities.} The error stems from failure 
    to perceive, interpret, or integrate information across 
    heterogeneous visual formats (text, tables, charts, figures, 
    and diagrams). Examples include misreading a table cell, 
    misinterpreting a chart axis, or failing to combine numerical 
    data from a table with surrounding textual context.
    
    \item \textbf{Structural Semantics.} The error stems from 
    failure to understand and navigate the document's logical 
    organization. Documents encode meaning through heading 
    hierarchies, cross-references, section flow, and spatial 
    layout. Examples include failing to follow a cross-reference 
    (``see Figure~5'' on page~3 pointing to page~8), or failing 
    at multi-hop reasoning along the document's structural chain.
    
    \item \textbf{Scale and Complexity.} The error stems from the 
    sheer volume of the document overwhelming the model. The 
    evidence is buried in a long document without distinctive 
    structural cues pointing to it (e.g., one ordinary paragraph 
    among hundreds). The challenge is pure scale, not structural 
    navigation or visual perception.
\end{itemize}

A key design choice is that we apply different 
classification strategies for each stage, reflecting the 
fundamentally different nature of retrieval and reasoning failures:

\begin{itemize}[nosep]
    \item \textbf{Fast Reading errors} are classified using a 
    binary prompt that chooses between \textit{Structural 
    Semantics} and \textit{Scale and Complexity} only. We exclude 
    \textit{Rich Modalities} because Fast Reading is a navigation 
    decision: the model decides \emph{which pages to retrieve} 
    before it perceives the visual content on those pages. If the 
    model fails to retrieve a page containing a chart, the failure 
    is not about the chart's modality but about the model's 
    inability to navigate to that page via structural cues or the 
    page being lost among too many candidates.
    
    \item \textbf{Focused Thinking errors} are classified using a 
    ternary prompt that chooses among all three characteristics. 
    Here, the model has already retrieved the relevant pages, so 
    failures can arise from misperceiving visual content 
    (\textit{Rich Modalities}), misunderstanding structural 
    relationships across retrieved pages (\textit{Structural 
    Semantics}), or being overwhelmed by information density 
    (\textit{Scale and Complexity}).
\end{itemize}

Both classification prompts receive the question, ground-truth 
answer, model prediction, retrieved pages, ground-truth evidence 
pages, total document page count, and the model's reasoning trace 
as context. Classification is performed by 
Llama-3.3-70B-Instruct-Turbo with greedy decoding.

\paragraph{Step 3: Complexity Grouping.}
To reveal how error patterns vary with task complexity, we group 
questions along two orthogonal dimensions derived from ground-truth 
metadata:

\begin{itemize}[nosep]
    \item \textbf{Page distribution:} \textit{Single-page} 
    (evidence spans $\leq 1$ page) vs.\ \textit{Multi-page} 
    (evidence spans $> 1$ pages).
    \item \textbf{Modality involvement:} \textit{Single-modal} 
    (evidence involves one source type) vs.\ \textit{Cross-modal} 
    (evidence involves $> 1$ source types, among pure-text, table, 
    figure, chart, and layout).
\end{itemize}

This yields four complexity groups: Single-page/Single-modal, 
Single-page/Cross-modal, Multi-page/Single-modal, and 
Multi-page/Cross-modal. For each group, we compute the proportion 
of questions falling into each stage $\times$ characteristic 
category, visualized in Figure~\ref{fig:error_decomposition}.

\subsection{Training Dynamics Metrics}
\label{appd:training_metrics}

In Section~\ref{sec: analysis}, we analyze the training dynamics of different 
strategies using two complementary metrics: self-certainty and 
perplexity.

\paragraph{Self-Certainty.}
We measure a model's confidence over its generated sequences using 
sequence-level self-certainty. Given a sequence of $n$ logit 
vectors $L = (l^{(1)}, \dots, l^{(n)})$, where 
$l^{(t)} = (l_1^{(t)}, \dots, l_K^{(t)})$ is the $K$-dimensional 
logit vector for the $t$-th token, self-certainty is defined as:
$$
\mathcal{S}_{\text{seq}}(L) = \frac{1}{n} \sum_{t=1}^{n}
\left[ \log \left( \sum_{i=1}^{K} e^{l_i^{(t)}} \right)
- \frac{1}{K} \sum_{i=1}^{K} l_i^{(t)} \right]
$$
This metric measures the average peakedness of the model's 
predictive distribution: a higher value indicates greater 
certainty over the generated sequence. In Figure~\ref{fig:unified_analysis_1}, we use this 
metric to show that the Direct RL policy maintains higher 
self-certainty than the SFT-initialized policies throughout 
training.

\paragraph{Perplexity.}
We use perplexity (PPL) to measure how well a model's learned 
distribution aligns with a given dataset. For a sequence of $n$ 
tokens $(x_1, \dots, x_n)$, perplexity under policy $\pi_\theta$ 
is defined as:
$$
\text{PPL} = \exp\!\left( -\frac{1}{n} \sum_{t=1}^{n} 
\log \pi_\theta(x_t \mid x_{<t}) \right)
$$
A lower PPL indicates that the model assigns higher probability 
to the observed tokens. In Figure~\ref{fig:unified_analysis_1}, we evaluate each 
training recipe's PPL on Fast Reading training data to diagnose 
inter-stage interference: if a Stage~2 SFT objective causes 
increasing PPL on Stage~1 data during training, this is consistent with the two objectives being antagonistic within 
the shared parameter space.

\section{Detailed Experimental Results}
\label{app:stage_level}

\subsection{Stage 1 Retrieval Performance Against RAG Baselines}
\label{app:retrieval}

Table~\ref{tab:main_results} reports end-to-end accuracy. This subsection reports the Stage 1 metrics that underlie it, for the main UniDoc configurations alongside the RAG baselines. We use the metrics defined in Section~\ref{sec: analysis}: Acc. is exact set match between the predicted and ground-truth evidence page sets, and Rec. indicates whether the predicted set fully contains the ground-truth set.

\begin{table}[h]
\caption{Stage 1 retrieval performance across three benchmarks, with end-to-end accuracy from Table~\ref{tab:main_results} for reference.}
\label{tab:retrieval_full}
\centering
\resizebox{\textwidth}{!}{
\begin{tabular}{l ccc ccc ccc}
\toprule
\multirow{2}{*}{\textbf{Method}} & \multicolumn{3}{c}{\textbf{MMLongBench-Doc}} & \multicolumn{3}{c}{\textbf{SlideVQA}} & \multicolumn{3}{c}{\textbf{DUDE}} \\
\cmidrule(r){2-4} \cmidrule(r){5-7} \cmidrule(r){8-10}
& Acc. & Rec. & E2E & Acc. & Rec. & E2E & Acc. & Rec. & E2E \\
\midrule
SV-RAG & 17.23 & 46.56 & 23.0 & 23.59 & 67.56 & 32.0 & 27.24 & 76.55 & 45.0 \\
FRAG & 3.70 & 54.81 & 31.3 & 4.97 & 71.92 & 52.4 & 23.81 & 78.10 & 43.1 \\
UniDoc-7B & 44.30 & 60.86 & 33.0 & 64.54 & 75.31 & 58.3 & 53.70 & 79.30 & 46.2 \\
UniDoc-8B & \textbf{51.64} & \textbf{72.32} & \textbf{52.5} & \textbf{70.26} & \textbf{80.59} & \textbf{65.1} & \textbf{56.65} & \textbf{83.55} & \textbf{47.8} \\
\bottomrule
\end{tabular}
}
\end{table}

UniDoc leads on both retrieval metrics across all three benchmarks, and the margin on exact set match is the larger of the two. This reflects a difference in retrieval formulation rather than in localization quality alone. The RAG baselines return a fixed number of pages per query regardless of how much evidence the question requires, so the returned set rarely coincides exactly with the ground-truth set even when it contains all of it. FRAG on MMLongBench-Doc illustrates this: recall reaches 54.81 while exact match is 3.70. Fast Reading instead predicts the evidence set itself, and its size adapts to the query.

\subsection{Performance Under Gold Evidence Pages}
\label{app:gold}

End-to-end accuracy conflates the quality of evidence localization with the quality of reasoning over that evidence. To isolate the second, we supply the ground-truth evidence pages directly and measure answer accuracy. The base models are evaluated within the same two-stage scaffold without our training, so the comparison holds the pipeline fixed and varies only the Stage 2 policy.

\begin{table}[h]
\caption{Accuracy given gold evidence pages, with end-to-end accuracy under the same two-stage scaffold for reference. DUDE is reported in ANLS.}
\label{tab:gold_evidence}
\centering
\begin{tabular}{l cc cc cc}
\toprule
\multirow{2}{*}{\textbf{Method}} & \multicolumn{2}{c}{\textbf{MMLongBench-Doc}} & \multicolumn{2}{c}{\textbf{SlideVQA}} & \multicolumn{2}{c}{\textbf{DUDE}} \\
\cmidrule(r){2-3} \cmidrule(r){4-5} \cmidrule(r){6-7}
& Gold & E2E & Gold & E2E & Gold & E2E \\
\midrule
Qwen2.5-VL-7B & 56.3 & 28.3 & 84.3 & 40.4 & 76.8 & 45.4 \\
UniDoc-7B & 72.1 & 33.0 & 86.2 & 58.3 & 80.5 & 46.2 \\
Qwen3-VL-8B & 75.6 & 43.4 & 88.1 & 52.6 & 77.6 & 46.7 \\
UniDoc-8B & \textbf{86.5} & \textbf{52.5} & \textbf{90.5} & \textbf{65.1} & \textbf{81.7} & \textbf{47.8} \\
\bottomrule
\end{tabular}
\end{table}

Given identical evidence, UniDoc outperforms its backbone on all three benchmarks. The margin is largest on MMLongBench-Doc, at 15.8 and 10.9 points for the 7B and 8B models, where evidence is distributed across long documents and Focused Thinking must synthesize across pages. On SlideVQA and DUDE it is smaller, at 1.9 and 2.4 and at 3.7 and 4.1 respectively, since the reasoning demanded once the evidence is in hand is correspondingly lighter and the base models already score highly under gold evidence. Focused Thinking therefore contributes independently of retrieval quality, and the end-to-end gains reported in Table~\ref{tab:main_results} come from both stages rather than from improved localization alone.

\subsection{Parameter Sharing Versus Decoupled Specialists}
\label{app:decoupled}

Section~\ref{sec:framework} executes both stages with a single set of shared parameters. This subsection compares that design against a decoupled alternative in which two independently trained models are composed sequentially at inference time. We evaluate three configurations that differ in whether parameters are shared and whether stage-level SFT is applied before RL. All three use the same training data and are evaluated end to end on Qwen3-VL-8B.

\begin{table}[h]
\caption{Shared parameters versus decoupled specialists, on Qwen3-VL-8B. Shared: both stages executed by one set of parameters (UniDoc-8B). Decoupled: two independently trained specialists composed sequentially at inference. SFT prior indicates whether stage-level SFT precedes RL. Params refers to the number of parameter sets held at inference time.}
\label{tab:decoupled}
\centering
\begin{tabular}{l c c ccc}
\toprule
\textbf{Configuration} & \textbf{SFT prior} & \textbf{Params} & \textbf{MMLong.} & \textbf{SlideVQA} & \textbf{DUDE} \\
\midrule
(1) Shared & no & 1$\times$ & 52.5 & 65.1 & 47.8 \\
(2) Decoupled & no & 2$\times$ & 48.1 & 59.4 & 48.6 \\
(3) Decoupled & yes & 2$\times$ & \textbf{60.4} & \textbf{74.8} & \textbf{66.7} \\
\bottomrule
\end{tabular}
\end{table}

Configuration (2) isolates the effect of parameter sharing under identical training. With no SFT prior, the shared-parameter model exceeds two independently trained specialists by 4.4 points on MMLongBench-Doc and 5.7 on SlideVQA, at half the inference-time parameter count.

Locating evidence and reasoning over it are both reasoning over the same document, and what separates them is the perceptual resolution at which that reasoning is performed. Under a shared policy the two stages draw on one body of reasoning ability, so structural reasoning learned while identifying evidence pages and fine-grained reasoning learned while interpreting them accumulate in the same parameters and can reinforce each other. Splitting them into two specialists forfeits this: each model learns its own stage in isolation, and the reasoning acquired by one is unavailable to the other. We attribute the gap in configuration (2) to that separation. It is largest on the long-context benchmarks, where the coupling between localization and reasoning is load-bearing; on DUDE, which averages 5.7 pages per question and leaves little for localization to do, the two configurations are comparable, with the decoupled pipeline 0.8 points ahead.

Configuration (3) adds the stage-level SFT prior to each specialist, and the ordering reverses. This is consistent with the optimization tension analyzed in Section~\ref{sec: analysis}: separate parameter sets can absorb both priors where a shared set cannot. Configuration (3) attains the strongest end-to-end accuracy on all three benchmarks, and it is built entirely from our contributions: the two specialists are trained on our synthesized data with the stage-level reasoning annotations, and they are composed within the Fast Reading and Focused Thinking framework. It therefore measures what our data and framework deliver when the constraint of a single parameter set is lifted. UniDoc reports Direct RL because it is the strongest configuration under a single set of parameters, which is the setting the framework targets, and configuration (3) reaches its accuracy at twice the parameters and two separate training pipelines.

\subsection{Retaining Low-Resolution Context in Focused Thinking}
\label{app:foveated}

In the design of Section~\ref{sec:framework}, a page omitted by Fast Reading is unavailable to Focused Thinking. This subsection reports a variant that removes this restriction. We retain the non-selected pages in the Focused Thinking context at $\phi_{\text{low}}$, so that Focused Thinking attends to the selected pages at $\phi_{\text{high}}$ alongside the remainder of the document at low fidelity, and no page leaves the context. The selected pages are re-encoded at $\phi_{\text{high}}$ as before; the non-selected pages remain at $\phi_{\text{low}}$, so their visual tokens are identical to those already produced during Fast Reading and are reused directly. Because both stages are executed by the same model with shared parameters, these tokens transfer without re-encoding, and the only cost is a longer Stage 2 context.

We partition MMLongBench-Doc by whether the page set returned by Fast Reading covers the ground-truth evidence pages, since the retained low-resolution context can only act on the questions where it does not.

\begin{table}[h]
\caption{Accuracy on MMLongBench-Doc, partitioned by whether the pages returned by Fast Reading contain every ground-truth evidence page. Under the proposed variant, a missing page is present at $\phi_{\text{low}}$.}
\label{tab:foveated}
\centering
\begin{tabular}{l c cc c cc}
\toprule
\multirow{2}{*}{\textbf{Setting}} & \multicolumn{3}{c}{\textbf{UniDoc-7B}} & \multicolumn{3}{c}{\textbf{UniDoc-8B}} \\
\cmidrule(r){2-4} \cmidrule(r){5-7}
& $n$ & Base & + low-res context & $n$ & Base & + low-res context \\
\midrule
Full evidence retrieved & 664 & 51.56 & 48.85 & 789 & 70.30 & 68.21 \\
Partial evidence retrieved & 427 & 4.24 & 24.66 & 302 & 6.03 & 36.97 \\
Overall & 1091 & 33.04 & \textbf{39.38} & 1091 & 52.51 & \textbf{59.56} \\
\bottomrule
\end{tabular}
\end{table}

The effect is concentrated where the mechanism is designed to act. On the partial-evidence partition, where the required page is available only at $\phi_{\text{low}}$, accuracy rises from 4.24 to 24.66 on UniDoc-7B and from 6.03 to 36.97 on UniDoc-8B, confirming that omitted evidence is recovered from the low-resolution context. On the full-evidence partition, where the retained pages carry no evidence the model needs, the additional context costs 2.71 and 2.09 points respectively. The net effect is positive on both backbones, with overall accuracy rising by 6.34 and 7.05 points.

\subsection{MP-DocVQA with In-Domain Training}
\label{app:mpdocvqa}

MP-DocVQA is reported as a held-out benchmark in Table~\ref{tab:main_results}: its training set is excluded from our training data, while SlideVQA and DUDE in that table are trained on. This subsection reports the matched setting, in which MP-DocVQA is included in training, so that the two settings can be compared directly.

We sample 4K instances from the MP-DocVQA~\citep{tito2023hierarchicalmultimodaltransformersmultipage} training set and process them through our data synthesis engine to produce stage-level reasoning annotations, following the treatment given to DUDE and SlideVQA in Section~\ref{sec: data engine}. Training otherwise follows the Direct RL recipe used for the main results, with no change to hyperparameters.

\begin{table}[h]
\caption{MP-DocVQA accuracy under held-out and in-domain training. Held-out excludes the MP-DocVQA training set from our training data and corresponds to the values reported in Table~\ref{tab:main_results}; trained includes 4K sampled instances from it.}
\label{tab:mpdocvqa_indomain}
\centering
\begin{tabular}{l c c c}
\toprule
\textbf{Backbone} & \textbf{Base model} & \textbf{Held-out} & \textbf{Trained} \\
\midrule
Qwen2.5-VL-7B & 57.3 & 75.0 & 81.3 \\
Qwen3-VL-8B & 72.8 & 81.4 & 89.6 \\
\bottomrule
\end{tabular}
\end{table}

With MP-DocVQA in training, UniDoc reaches 81.3 and 89.6 on the two backbones, improving over the base models by 24.0 and 16.8 points. The held-out results, 75.0 and 81.4, remain well above the base models, so the framework transfers to a document type and query distribution absent from training, and the gap between the two settings reflects the absence of this distribution from training rather than a property of the method on this benchmark.

\section{Prompt Lists}

\subsection{LLM-Judge in Reinforcement Learning}
Please refer to Table \ref{appd: llmjudge prompt}.

\begin{tcolorbox}[colframe=black!75!white, colback=white, coltitle=white, title=Prompts for LLM-Judge in Reward Allocation, fonttitle=\bfseries, breakable]
\small
You are an impartial AI judge. Your goal is to evaluate the semantic similarity between a ``Predicted Answer'' and a ``Ground Truth Answer".

\noindent\makebox[\linewidth]{\tikz[baseline]{\draw[dashed] (0,0) -- (\linewidth,0);}}\vspace{2mm}
\vspace{0.1cm}
Question: \texttt{\{question\}}

\vspace{0.1cm}
Ground Truth Answer: \texttt{\{ground truth\}}

\vspace{0.1cm}
Model Response: \texttt{\{prediction\}} \\
\noindent\makebox[\linewidth]{\tikz[baseline]{\draw[dashed] (0,0) -- (\linewidth,0);}}\vspace{2mm}
Based on the context, compare the predicted answer to the ground truth and determine a score based on the following guideline:
\begin{itemize}[leftmargin=*]
    \item First, recover a reference answer sentence based on the question context and the ground truth answer. Write ``The reference answer sentence is: [sentence here based on the question and ground truth]"
    \item Second, what is the answer sentence implied by the model response? Write ``The model implies the answer sentence: [sentence here implied from the model response]"
    \item Third, compare and reason about the two sentences, are they delivering the same meanings? Write ``Analysis and Conclusion"
        \begin{itemize}
            \item For Numerical Ground Truth Answers: If the ground truth is numerical values, such as integers, floats, or something like math formula -- you need to check whether the model response has an exact match with the numerical values. The score is 1.0 for an exact match and 0.0 otherwise. Account for units like ``million". E.g., 9.5 million should be equivalent to 9,500,000
            \item For Textual Answers: Assign a semantic similarity score from 0.0 (no overlap) to 1.0 (perfect equivalence). A verbatim match is not required for a score of 1.0. But give the perfect score 1.0 sparingly, only when the answers are really identical in meaning. 
        \end{itemize}
    \item If the Predicted Answer use NO boxed notation to properly highlight the key answers, reduce the assigned score by 0.3 (but the resulting score is no less than 0)
\end{itemize}

\textbf{Output:}
Follow the above guideline to provide the analysis of your comparison. Then, on a new line, write the numeric score enclosed in boxed.

\end{tcolorbox}
\captionof{table}{Prompts for LLM-Judge in Reinforcement Learning.}

\label{appd: llmjudge prompt}

\subsection{Data Engine Generator Prompt}
Please refer to Table \ref{appd: data engine generator}.

\begin{tcolorbox}[colframe=black!75!white, colback=white, coltitle=white, title=Prompts for Query Generator, fonttitle=\bfseries, breakable]
\small

\textbf{Role and Goal:}

You are an expert instructional designer, tasked with creating a high-quality question-answer pair based on a provided image, its accompanying text and a set of keypoints. Your goal is to assess a student's ability to synthesize information from visual and textual sources and their ability to perform REASONING with synthesized information.

Specifically, you will be given three kinds of text snippets:
\vspace{0.1cm}
\begin{itemize}[leftmargin=*]
    \item A subtitle that describes the image.
    \item A related text snippet that provides additional context or information about the image.
    \item A set of key points extracted from the image and text which are core ideas or findings of the image and text.
\end{itemize}

\vspace{0.3cm}

\textbf{Input:}
\vspace{0.1cm}

Subtitle of the image: \texttt{\{subtitle\}}

Related text snippet: \texttt{\{related text\}}

Set of key points extracted from the image and text: \texttt{\{key points\}}

\vspace{0.3cm}

\textbf{Critical Instructions:}

\vspace{0.1cm}
1.  Question Crafting:
\vspace{0.1cm}
\begin{itemize}[leftmargin=*]
    \item The question MUST be answerable ONLY by using the information present in BOTH the image and the texts.
    \item The question should require the student to synthesize information from the image, the subtitle and the related text. All three sources MUST be used to answer the question.
    \item The question should test the student's ability from two steps: 1. Attention to detail: Firstly, the student should observe the image and the texts to gather relevant information. 2. Reasoning: Secondly, the student should use the gathered information to answer the question.
    \item The keypoints are provided to give you direction on what kind of information is important in the image and texts, so you can use them to generate a more focused question. It's preferrred to combine multiple key points to generate a challenging question and generate a question that requires multiple steps of reasoning.
    \item The question must be clear, concise, and unambiguous. The question must not be subjective or open-ended; it should have a definitive answer based on the provided materials. For example, it should not ask ``What do you think about the image?'' or ``What is your opinion on the text?". Instead, it should ask for specific information that can be found in the image and the texts.
    \item There should be **only** one ``what/how/why'' question in the output, you must not generate multiple questions, which means outputs like ``What ... and how ...?'' or ``Why...? How...?'' is strictly banned, because they contained two questions instead of one. 
\end{itemize}

\vspace{0.1cm}
2.  The Golden Rule of Answering:
\vspace{0.1cm}
\begin{itemize}[leftmargin=*]
    \item The provided answer MUST be derived exclusively from the given materials.
    \item Do NOT use any external knowledge, assumptions, or logical inferences that are not directly supported by the provided content. The answer should be a verifiable fact from the source.
    \item The answer should to be the ultimate result of a reasoning process using the key points, related text, subtitle and image, rather than just a simple fact from the image or text. The key points are used to provide necessary context or information.
    \item The answer should be within a few words or a short sentence, but NOT single number or single word. It should directly providing the answer without additional explanation or context.
\end{itemize}
\end{tcolorbox}
\captionof{table}{Prompts for query generator in data engine.}

\label{appd: data engine generator}

\subsection{Data Engine Filterer Prompt}
Please refer to Table \ref{appd: data engine filterer}.

\begin{tcolorbox}[colframe=black!75!white, colback=white, coltitle=white, title=Prompts for filterer in data engine, fonttitle=\bfseries]
\small
You are a helpful assistant that evaluates the correctness of answers based on given questions and reference answers.

\textbf{You are given the following information:}

\begin{itemize}[leftmargin=*]
    \item A question
    \item A reference answer
    \item Some images which can provides information to solve the question
\end{itemize}

\textbf{Your task is listed below:}

\begin{itemize}[leftmargin=*]
    \item First check if the question is solvable with the given images. The unsolvable questions include:
        \begin{itemize}[leftmargin=*]
            \item Questions that require information not present in the images.
            \item Questions that are too vague or ambiguous to answer based on the images or there are multiple interpretations.
            \item Questions that require subjective interpretation or opinion that cannot be derived from the images.
        \end{itemize}
    The solvable questions include:
        \begin{itemize}[leftmargin=*]
            \item Questions that can be answered directly from the images.
            \item Questions that can be answered with a reasonable inference based on the images.
        \end{itemize}
    \item If the question is solvable, check if the reference answer is correct.
    \item Some images which can provides information to solve the question. If the question is solvable and the reference answer is correct, return ``solvable", the reference answer and a brief explanation of why the answer is correct. If the question is solvable but the reference answer is incorrect, return ``solvable", a corrected answer and a brief explanation of the correction. If the question is not solvable, return ``unsolvable", a brief explanation of why it cannot be answered with the given information.
\end{itemize}

\textbf{Important Notes:}
\begin{itemize}[leftmargin=*]
    \item The corrected answer should be concise and directly address the question, avoiding unnecessary details.
    \item The explanation should be clear and concise, providing enough context to understand the reasoning without being overly verbose.
    \item The response should be in JSON format as specified above.
\end{itemize}

\noindent\makebox[\linewidth]{\tikz[baseline]{\draw[dashed] (0,0) -- (\linewidth,0);}}\vspace{2mm}

User Query: \texttt{\{question\}}

Reference Answer: \texttt{\{reference answer\}}

\end{tcolorbox}
\captionof{table}{Prompts for Filterer in data engine.}

\label{appd: data engine filterer}

\section{Case Study}
We present qualitative case studies from two perspectives. First, 
we examine the queries generated by our Data Synthesis Engine, 
demonstrating their \textbf{spatially distributed and complex 
nature}. This validates that our data is explicitly tailored to 
challenge the evidence retrieval and multimodal reasoning 
capabilities required by Fast Reading and Focused Thinking. 
Second, we conduct a comparative analysis of inference 
trajectories before and after SFT with our synthesized 
stage-level reasoning annotations, empirically validating their 
effectiveness.

\subsection{Query Analysis}
\begin{figure*}
  \centering
  \includegraphics[width=0.80\textwidth]{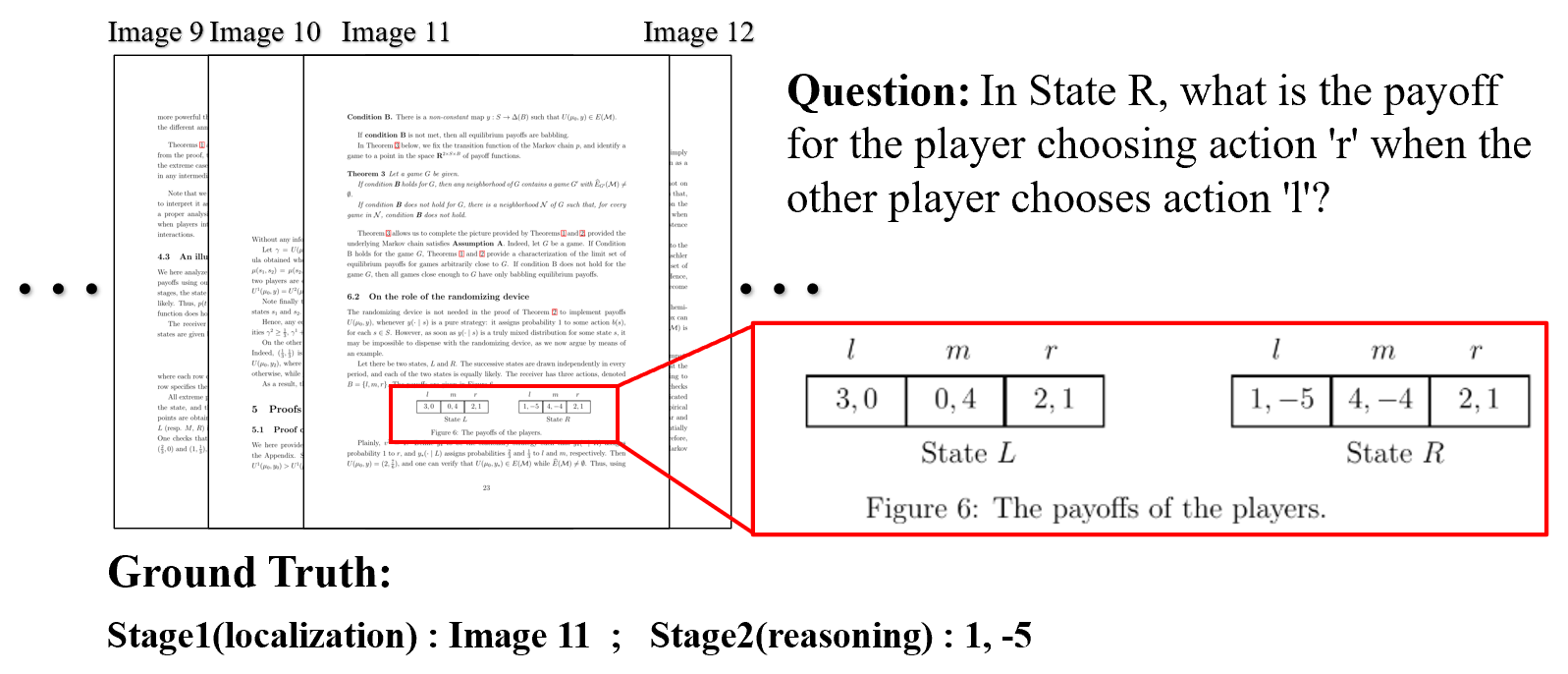}
  \caption{A synthesized query example. The query provides 
  training signals for both stages: Fast Reading must retrieve 
  the page containing the multimodal anchor ``Figure 6'', while 
  Focused Thinking must reason over ``Figure 6'' and its 
  referencing text to derive the answer.}
  \label{fig:query_sample}
\end{figure*}

\begin{figure*}
  \centering
  \includegraphics[width=0.80\textwidth]{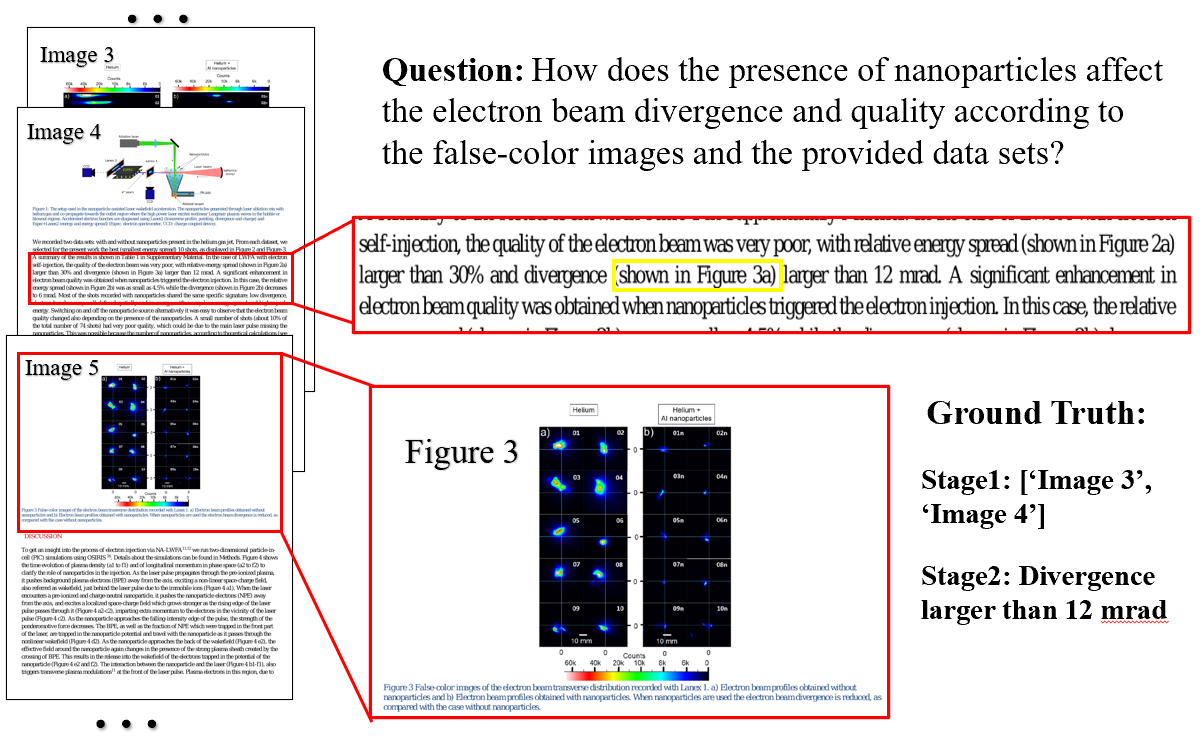}
  \caption{A synthesized query requiring multi-page evidence. 
  Fast Reading must retrieve both the page containing ``Figure 3'' 
  and the page referencing it with deeper analysis. Focused 
  Thinking must then synthesize information across the retrieved 
  pages to produce the answer.}
  \label{fig:query_sample1}
\end{figure*}

We present representative training samples generated by our 
synthesis engine in Figure~\ref{fig:query_sample} and 
Figure~\ref{fig:query_sample1}. Given a document $d$, the engine 
outputs a training triplet consisting of the query $q$, the 
ground-truth answer $y$, and the evidence page set 
$\mathcal{P}_{\text{gt}}$ required to derive that answer.

This triplet provides distinct training signals for both stages:
\begin{itemize}[nosep]
    \item \textbf{Fast Reading:} The full document $d$, the query 
    $q$, and the evidence pages $\mathcal{P}_{\text{gt}}$ serve 
    as the training signal. The model must learn to reason over 
    the document's structure to retrieve the correct pages.
    \item \textbf{Focused Thinking:} The high-resolution content 
    of $\mathcal{P}_{\text{gt}}$, along with $q$ and $a$, serves 
    as the training signal. The model must perform fine-grained 
    multimodal reasoning to extract the final answer.
\end{itemize}

In the example shown in Figure~\ref{fig:query_sample}, the query 
targets a visual anchor (``Figure 6: The payoffs of the 
players'') located on Image~11. The model first performs Fast 
Reading to retrieve this page from the full document. It then 
transitions to Focused Thinking, performing multimodal reasoning 
over the payoff matrix in Figure~6 and its referencing text, 
specifically aligning ``State R'', the player's action ``r'', 
and the opponent's action ``l'', to derive the final answer 
``1, -5''.

Figure~\ref{fig:query_sample1} illustrates a more complex 
scenario where the query requires retrieving \textit{multiple} 
disjoint pages. This forces the model to perform holistic 
structural reasoning, simultaneously identifying and aggregating 
evidence distributed across different parts of the document. 
This example demonstrates the spatially distributed nature of 
queries produced by our engine, requiring the model to understand 
document-wide cross-references rather than relying on local 
context.

\subsection{Qualitative Analysis of Reasoning Annotations}
We present a qualitative comparison between the model fine-tuned 
with our stage-level reasoning annotations and the base model. 
Our analysis validates two design objectives: (1) Fast Reading 
annotations endow the model with \textbf{structure-aware} 
retrieval capabilities, enabling it to determine page relevance 
through holistic reasoning rather than keyword matching; (2) 
Focused Thinking annotations facilitate an \textbf{adaptive 
grounding} mechanism, allowing the model to dynamically adjust 
its reasoning depth to queries of varying visual complexity.

\begin{figure*}
  \centering
  \includegraphics[width=0.80\textwidth]{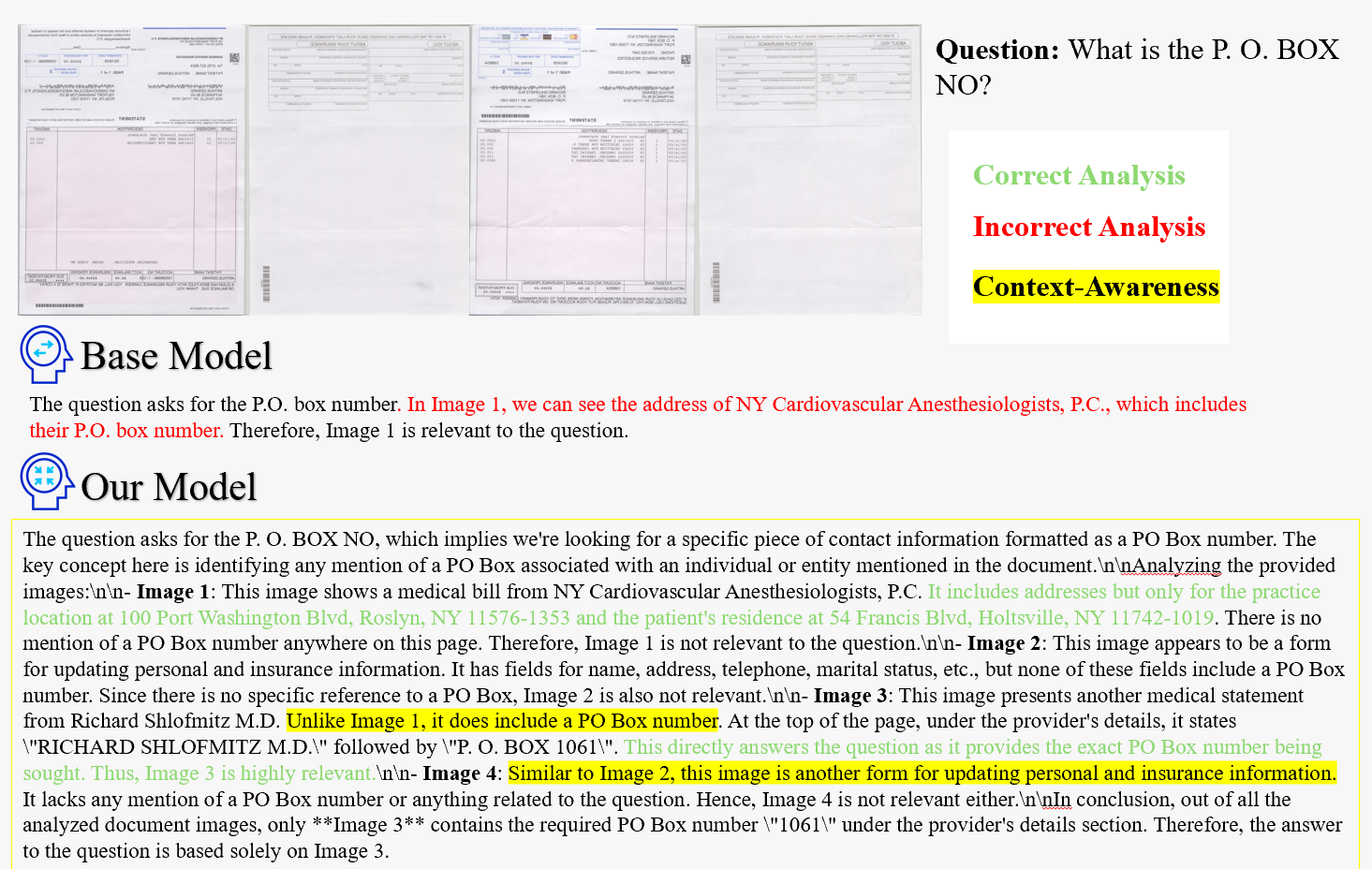}
  \caption{Fast Reading comparison. The base model fails by 
  relying on keyword matching without holistic context. Our model 
  cross-references pages to correctly retrieve the evidence.}
  \label{fig:stage1_sample}
\end{figure*}

In Figure~\ref{fig:stage1_sample}, we compare the base model and 
our model on a Fast Reading task. The base model attempts to 
provide a direct answer without explicit reasoning; while it 
successfully attends to keywords like ``P.O.\ box number'', it 
fails to incorporate context from subsequent pages, leading to a 
conceptual misunderstanding. In contrast, our model employs a 
structure-aware strategy: while analyzing pages, it explicitly 
cross-references the current page with others to verify semantic 
continuity. This holistic reasoning enables it to correctly 
retrieve the relevant pages.

\begin{figure*}
  \centering
  \includegraphics[width=1.00\textwidth]{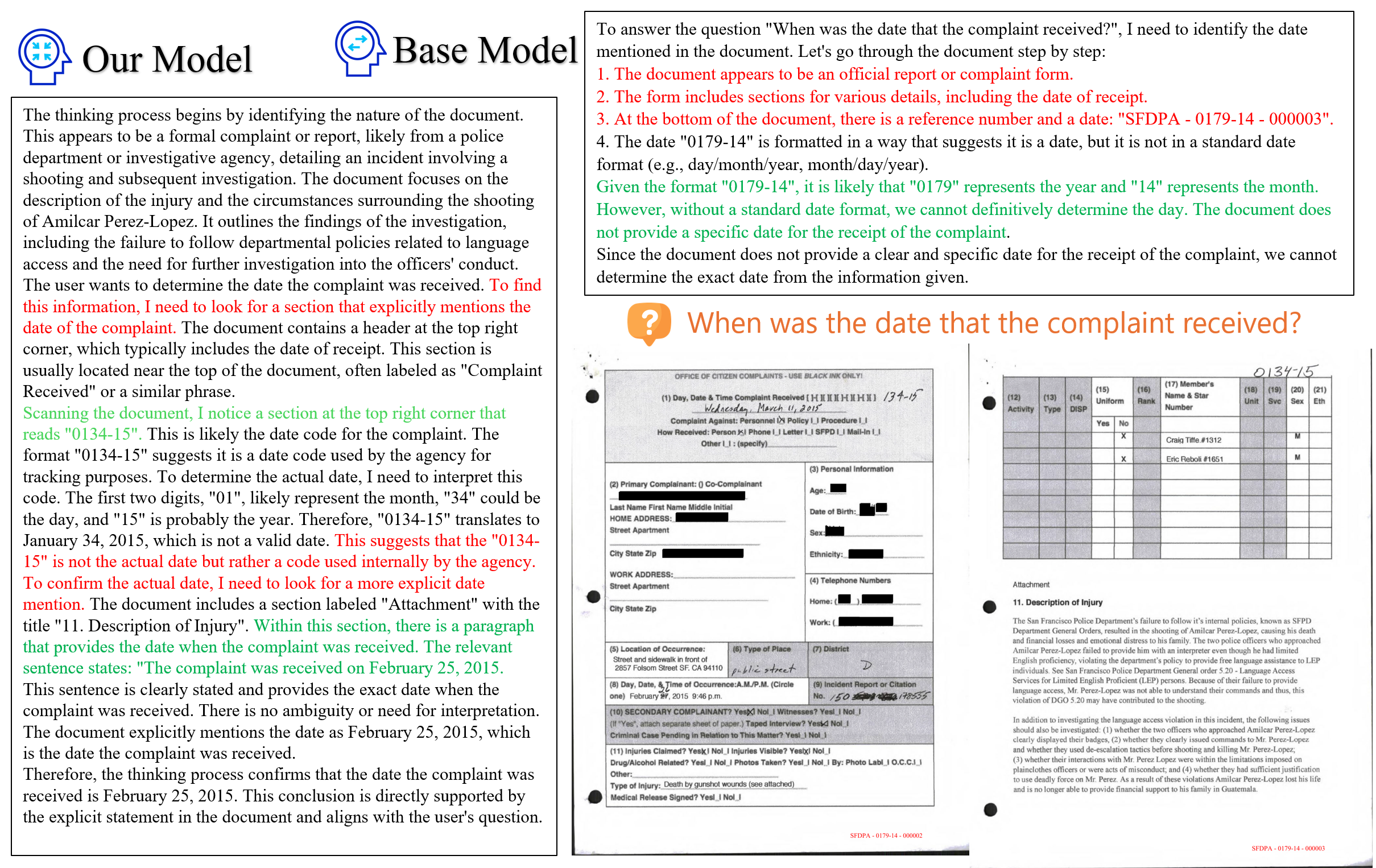}
  \caption{Focused Thinking comparison. \textcolor{red}{Red text} 
  indicates grounding steps and \textcolor{green}{green text} 
  indicates extraction steps. The base model halts after the 
  first failed attempt. Our model iteratively re-grounds and 
  extracts until sufficient evidence is gathered.}
  \label{fig:stage2_sample}
\end{figure*}

Figure~\ref{fig:stage2_sample} demonstrates the efficacy of our 
Focused Thinking annotations. The base model fails because it 
halts when the initial grounding attempt yields insufficient 
information. Our model, guided by the annotations, employs an 
iterative approach: it constructs a chain of grounding and 
extraction steps, re-grounding to locate additional evidence when 
the first attempt is inconclusive, ultimately arriving at the 
correct answer.

\end{document}